\documentclass[final]{cvpr}

\usepackage{times}
\usepackage{epsfig}
\usepackage{graphicx}
\usepackage{amsmath}
\usepackage{amssymb}

\usepackage[pagebackref=true,breaklinks=true,colorlinks,bookmarks=false]{hyperref}

\def\cvprPaperID{2234} 
\def\confYear{CVPR 2021}

\usepackage{multirow}
\usepackage{amsmath,amssymb} 
\usepackage{xcolor}
\usepackage{tabularx}
\usepackage{arydshln}
\usepackage[labelformat=simple]{subcaption}

\usepackage[switch]{lineno}

\usepackage{pythonhighlight}
\usepackage{pdflscape}
\usepackage{rotating}

\newcommand{\bsf}[1]{\textbf{\textsf{#1}}}
\newcommand{\tb}[1]{\textbf{#1}}

\newcommand{\lxm}[1]{\textcolor[rgb]{1,0,1}{#1}}
\newcommand{\ylb}[1]{\textcolor[rgb]{0,0,1}{#1}}

\renewcommand{\lxm}[1]{{#1}}
\renewcommand{\ylb}[1]{{#1}}

\renewcommand{\paragraph}[1]{{\noindent \textbf{#1}}}

\DeclareMathOperator*{\argmin}{arg\,min}

\newcolumntype{C}{>{\centering\arraybackslash}X}
\newcolumntype{L}{>{\raggedright\arraybackslash}X}
\newcolumntype{R}{>{\raggedleft\arraybackslash}X}
\newcommand\blankfootnote[1]{%
  \let\svthefootnote\thefootnote%
  \let\thefootnote\relax\footnotetext{#1}%
  \let\thefootnote\svthefootnote%
}

\begin{document}

\title{Progressive Semantic-Aware Style Transformation for Blind Face Restoration}

\author {
    Chaofeng Chen\textsuperscript{\rm 1, 3$*$} \quad
    Xiaoming Li\textsuperscript{\rm 2, 3, 5} \quad
    Lingbo Yang\textsuperscript{\rm 4, 3} \quad
    Xianhui Lin\textsuperscript{\rm 3}  \\
    Lei Zhang\textsuperscript{\rm 3, 5} \quad 
    Kwan-Yee K. Wong \textsuperscript{\rm 1} \\
    
    \small\textsuperscript{\rm 1} Department of Computer Science, The University of Hong Kong \quad 
    \small\textsuperscript{\rm 2} Faculty of Computing, Harbin Institute of Technology \\
    \small\textsuperscript{\rm 3} DAMO Academy, Alibaba Group \quad
    \small\textsuperscript{\rm 4} Institute of Digital Media, Peking University \\
    \small\textsuperscript{\rm 5} Department of Computing, The Hong Kong Polytechnic University \\
    \tt\small\url{https://github.com/chaofengc/PSFRGAN}
}

\twocolumn[{%
\renewcommand\twocolumn[1][]{#1}%
\maketitle
\begin{center}
    \centering
    \includegraphics[width=.99\textwidth]{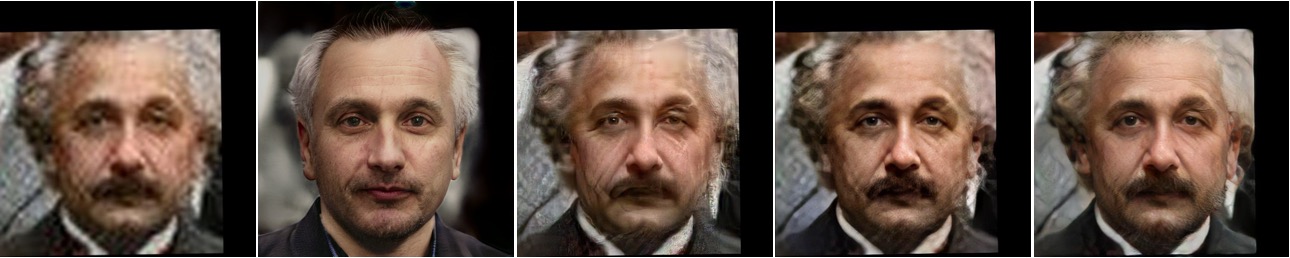}
    \begin{tabularx}{.99\linewidth}{*{5}{C}}
    LQ & PULSE \cite{menon2020pulse} & HiFaceGAN \cite{Yang2020HiFaceGANFR} & DFDNet \cite{Li_2020_ECCV} & \tb{Ours} 
    \end{tabularx}
    \captionof{figure}{Result comparison with state-of-the-art methods. (Please zoom in to see the details)}
\end{center}%
}]

\maketitle

\blankfootnote{$^*$ This work was done when he was an intern at Alibaba.}

\begin{abstract}
Face restoration is important in face image processing, and has been widely studied in recent years. However, previous works often fail to generate plausible high quality (HQ) results for real-world low quality (LQ) face images. In this paper, we propose a new progressive semantic-aware style transformation framework, named PSFR-GAN, for face restoration. Specifically, instead of using an encoder-decoder framework as previous methods, we formulate the restoration of LQ face images as a multi-scale progressive restoration procedure through semantic-aware style transformation. Given a pair of LQ face image and its corresponding parsing map, 
we first generate a multi-scale pyramid of the inputs, and then progressively modulate different scale features from coarse-to-fine in a semantic-aware style transfer way. Compared with previous networks, the proposed PSFR-GAN makes full use of the semantic (parsing maps) and pixel (LQ images) space information from different scales of input pairs. In addition, we further introduce a {\em semantic aware style loss} which calculates the feature style loss for each semantic region individually to improve the details of face textures. Finally, we pretrain a face parsing network which can generate decent parsing maps from real-world LQ face images. Experiment results show that our model trained with synthetic data can not only produce more realistic high-resolution results for synthetic LQ inputs but also generalize better to natural LQ face images compared with state-of-the-art methods.  
\end{abstract}

\section{Introduction}

\ylb{Blind face restoration} refers to recovering the HQ images from the LQ inputs which suffer from unknown degradation such as low resolution, noise, blur and lossy compression. It has drawn more and more interest due to its wide applications. However, most current restoration methods still focus on a specific type of restoration, especially super resolution, and few of them can generalize well to real LQ images. 

Unlike general image restoration, face restoration can exploit strong prior knowledge of the face to recover details of the face components even when the images are severely degraded. Therefore, many recent works about face super resolution incorporate face prior knowledge to improve the performance, such as parsing maps \cite{Chen2018CVPR}, face landmarks \cite{bulat2017super,progressive-face-sr,yu2018face} and identity prior \cite{zhang2018super}. Most of these works are based on an encoder-decoder like structure, \lxm{which follow} the practice of general image restoration and aim to learn a direct black-box mapping from LQ to HQ images. Although they managed to get better results with extra face prior knowledge as inputs or supervision, few of them reports satisfactory results for real LQ images. Other methods \cite{Li_2018_ECCV,Li_2020_CVPR} try to utilize high quality references to facilitate the restoration of LQ images. However, their practical applications are limited when there are no high quality references.

In this work, we propose a new progressive framework, named PSFR-GAN, which formulates face restoration as multi-scale semantic-aware style transformation procedure. Inspired by recent success of style based GAN, \ie, StyleGAN \cite{karras2019style}, we use a semantic-aware style transfer approach to modulate the features of different scales progressively. To be specific, the proposed PSFR-GAN starts with a learned constant latent code and then generates features of different scales through several upsample layers. We modulate the ``styles'' of different scale features by generating the corresponding style transformation parameters from different scale inputs. The LQ \lxm{input} provides color information and the parsing map provides shape and semantic information. In this way, more details are added to the final features in a coarse-to-fine manner. In addition, we proposed a {\em semantic aware style loss} which calculates the gram matrix loss for each semantic region separately. Gram matrix loss is usually applied in neural style transfer \cite{gatys2016neuralstyle}, recent works \cite{gondal2018unreasonable} found it also effective in recovering textures. In this work, we show that the {\em semantic aware style loss} can help to improve the restoration of textures and alleviate the occurrence of artifacts in different face regions. 

Finally, to make our framework more practical, we pretrain a face parsing network (FPN) for LQ face images. Intuitively, predicting face parsing maps is easier than face restoration because we do not need to care the texture details. Experiments demonstrate that the FPN is pretty robust on parsing real-world \lxm{LQ} face images. During test time, we first generate parsing \lxm{maps} for LQ inputs with FPN, and then produce the HQ outputs with PSFR-GAN. 

Our contributions are summarized as follows: 
\begin{enumerate}
  \item We propose a novel multi-scale progressive framework for practical blind face restoration, \ie PSFR-GAN. Our model can recover high quality face details progressively through semantic aware style transformation with multi-scale LQ images and parsing maps as inputs. Compared with previous works, the proposed PSFR-GAN can make better use of multi-scale inputs in both pixel domain and semantic domain. 
  \item We introduce the {\em semantic aware style loss} which helps to improve the texture restoration of different semantic regions and reduce the occurrence of artifacts.
  \item Extensive experiments demonstrate that our model trained with synthetic dataset generalizes better to natural LQ images than current state-of-the-arts. 
  \item By introducing a pretrained LQ face parsing network, our model can generate HQ images given only LQ inputs, making it highly practical and applicable. 
\end{enumerate}

\section{Related Works}

In this section, we briefly review \lxm{the} existing methods for face super-resolution, blind face restoration and HQ face generation based on generative adversarial networks.

\paragraph{Face Super-Resolution.} 
Face super resolution is often studied as a basic task for face restoration. Zhu \etal \cite{zhu2016deep} proposed a cascaded two-branch network to optimize face hallucination and dense correspondence field estimation in a unified framework. Yu \etal \cite{yu2016ultra} exploited generative adversarial networks (GAN) to directly super-resolve LR inputs. They further improved their model to handle unaligned faces \cite{yu2017face}, noisy faces \cite{yu2017hallucinating} and faces with different attributes \cite{yusuper}. Instead of directly inferring HR face images, Huang \etal \cite{huang2017wavelet} proposed to predict wavelet coefficients from LR images to reconstruct HR images. 
More recent works exploited extra face prior information to improve the SR performance. Chen \etal \cite{Chen2018CVPR} used face landmark heatmaps and parsing maps to super-resolve unaligned LR faces. They first predicted landmark heatmaps and parsing maps from LR faces, and then concatenated them with feature maps to fine tune the SR results. Adrian \etal \cite{bulat2017super} jointly learned face SR and landmark prediction. Yu \etal \cite{yu2018face} employed face component heatmaps to preserve face structure while super-resolving LR faces. They used one CNN branch to predict face component heatmaps and then concatenated the heatmaps to the SR CNN branch. Kim \etal \cite{progressive-face-sr} proposed a multi-scale facial attention loss by multiplying the heatmap values to the pixel differences of different scales to better restore pixels around landmarks. \lxm{However, most of these works followed a general encoder-decoder framework and did not fully utilize the face prior, making them unsuitable to handle real-world LQ images.}

\paragraph{Blind Face Restoration} 
To deal with blind face restoration, recent works either tried to improve the framework or adopted reference based approach. Adrian \etal \cite{bulatyang2018learn} proposed a two-stage GAN framework to learn real degradation. Yang \etal \cite{Yang2020HiFaceGANFR} introduced the HiFaceGAN which progressively replenish face details. As for reference based methods, Song \etal \cite{song-ijcai17-faceSR} constructed a high-resolution dictionary to help enhance face components. Li \etal \cite{Li_2018_ECCV} proposed GRFNet which learns to warp a guidance image for blind face restoration. They further improve their work by replacing single reference with multiple reference images \cite{Li_2020_CVPR} and feature dictionaries \cite{Li_2020_ECCV}. The aforementioned model based methods fail to make full use of face prior, and the reference based approaches require HQ references which limits their practical application.

\paragraph{High Quality Face Generation with GAN.} 
Among all GAN based image generation methods, the style based \lxm{GANs} (StyleGAN) proposed by Karras \etal \cite{karras2019style,karras2019analyzing} \lxm{exceed} other works by a large margin and generate HQ faces which are almost indistinguishable from real photos. Instead of directly generating the faces from random latent vector, they first generate style transform parameters from latent vector, and modulated the features with adaptive instance normalization (AdaIN) \cite{huang2017adain}. SEAN \cite{Zhu_2020_CVPR} and GauGAN \cite{park2019SPADE} extend StyleGAN to generate realistic faces from parsing maps. PULSE \cite{menon2020pulse} describes a self-supervised face restoration approach by exploring the latent space of a pretrained StyleGAN. mGANprior \cite{mganprior} extended PULSE with multiple latent codes. Inspired by the above works, we introduce the style transformation approach to face restoration in this paper. Compared with PULSE, our model does not need a time consuming optimization process and preserves the shape and identity better. 

\section{Proposed Method}

\begin{figure*}[htb]
  \includegraphics[width=.99\linewidth]{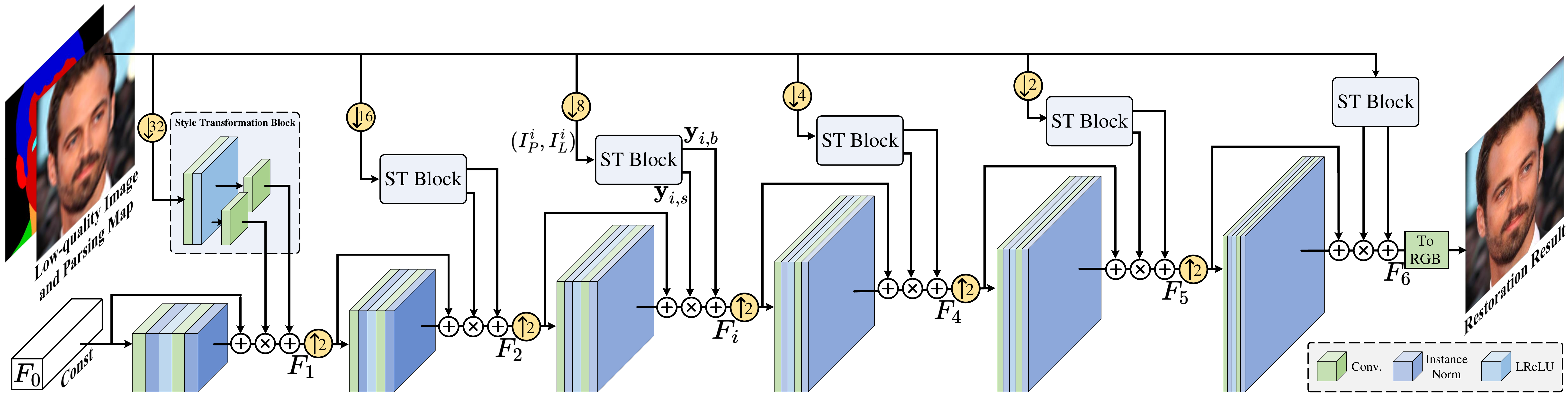}
  \caption{Visulization of the proposed progressive semantic-aware style transformation network for face restoration.}
  \label{fig:psfr-framework}
\end{figure*}

In this section, we first describe our formulation and framework in detail, then introduce the {\em semantic-aware style loss} and the other objectives used to train our networks. 

\subsection{Progressive Semantic-Aware Style Transformation}
As illustrated in Fig. \ref{fig:psfr-framework}, the architecture of our PSFR-GAN is inspired by GauGAN \cite{park2019SPADE} and StyleGANs \cite{karras2019style,karras2019analyzing}. It starts with a learned constant of size $C\times16\times16$, denoted as $F_0$, where $C$ is the channel size. Then, $F_0$ goes through several upsample residual blocks and generates the final features with the same size as HQ images. Let's define the output features of $i$-th residual block as $F_i$, then the features are progressively upsampled in the following way 
\begin{align}
    F_i = 
    \begin{cases}
        \Phi_{ST}\Bigl(\Phi_{RES}(F_{i-1})\Bigr), & i = 1\\
        \Phi_{ST}\Bigl( \Phi_{UP}(F_{i-1}) \Bigr), & 1 < i \leq 6 
    \end{cases}
\end{align}
where $\Phi_{RES}(\cdot)$ denotes the residual convolution block, $\Phi_{UP}(\cdot)$ denotes the upsample residual convolution block and $\Phi_{ST}(\cdot)$ denotes the style transformation block. The last feature $F_6$ goes through a single \texttt{ToRGB} convolution layer and predicts the final output $\hat{I_H}$. 

The  $\Phi_{ST}(\cdot)$ blocks are the key parts of our framework. Each of them learns the style transformation parameters $\mathbf{y}_i=(\mathbf{y}_{s,i}, \mathbf{y}_{b,i})$ for $F_i$ from corresponding scale of input pairs, \ie, LQ images and parsing maps denoted as $(I_L^i, I_P^i)$.  $(I_L^i, I_P^i)$ are resized to the same size as $F_i$ through bicubic interpolation. Then, $\Phi_{ST}(\cdot)$ can be formulated as follows
\begin{gather}
    \mathbf{y}_i = \Psi(I_L^i, I_P^i), \\
    F_i = \mathbf{y}_{i, s} \frac{\Phi_{UP}(F_{i-1}) - \mu\Bigl(\Phi_{UP}(F_{i-1})\Bigr)}{\sigma\Bigl(\Phi_{UP}(F_{i-1})\Bigr)} + \mathbf{y}_{i, b}, 1 < i \leq 6 
\end{gather}
where $\Psi(\cdot)$ is a lightweight network composed of several convolution layers. $\mu(\cdot)$ and $\sigma(\cdot)$ are the mean and standard variation of features. Compared with StyleGAN which adpots spatially invariant styles, we follow \cite{gu2019ikc,wang2018sftgan} and compute the spatially adaptive style parameters $\mathbf{y}_i$ with the same size as $F_i$. This helps to make full use of the spatial-wise color and texture information from $I_L$ as well as shape and semantic guidance from $I_P$. We use the pretrained face parsing network (FPN) to generate $I_P$ from LQ inputs $I_L$.   

\subsection{Semantic-Aware Style Loss} \label{sec:method-ss}
Recent super-resolution work \cite{gondal2018unreasonable} has demonstrated that gram matrix loss which is usually used in style transfer helps a lot in recovering textures. To better synthesize texture details, we introduce the {\em semantic-aware style loss} $\mathcal{L}_{ss}$ which calculates the gram matrix loss for each semantic region separately. We use VGG19 features of layer \texttt{relu3\_1}, \texttt{relu4\_1} and \texttt{relu5\_1} to calculate $\mathcal{L}_{ss}$. Denote $\phi_i$ as the $i$-th layer feature in VGG19 and the parsing mask with label $j$ as $M_j$ (the background is denoted as $M_0$), the semantic aware style loss is formulated as
\begin{equation}
   \mathcal{L}_{ss} = \sum_{i=3}^5 \sum_{j=0}^{18} \|\mathcal{G}\bigl(\phi_i(\hat{I_H}), M_j\bigr) - \mathcal{G}\bigl(\phi_i(I_H), M_j\bigr)\|_2,
\end{equation}   
$\mathcal{G}(\cdot)$ computes the gram matrix of feature $\phi_i(\cdot)$ with semantic label mask $M_j$ as below
\begin{equation}
    \mathcal{G}(\phi_i, M_j) = \frac{\bigl(\phi_i \odot M_j)^T \bigl(\phi_i \odot M_j)}{\sum M_j + \epsilon},
\end{equation}
where $\odot$ is the element-wise product, and $\epsilon=1e-8$ is used to avoid zero division. 

\subsection{Model Objectives} \label{sec:method-loss}

Following previous image restoration works \cite{Chen2018CVPR,wang2018esrgan,wang2018pix2pixHD,Li_2020_CVPR,Li_2020_ECCV}, we apply reconstruction loss and  adversarial loss apart from $L_{ss}$.  

\paragraph{Reconstruction Loss.} It is the combination of pixel and feature space mean square error (MSE) which aims to constrain the network output $\hat{I_H}$ close to ground truth $I_H$. We formulate the reconstruction loss as
\begin{equation}
    \mathcal{L}_{rec} = \|\hat{I_H} - I_H\|_2 + \sum_s \sum_{k=1}^4 \|D_s^k(\hat{I_H^s}) - D_s^k(I_H^s)\|_2
\end{equation}
The second term in $\mathcal{L}_{rec}$ is the multi-scale feature matching loss \cite{wang2018pix2pixHD} which matches the discriminator features of $\hat{I_H}$ and $I_H$. $s\in \{1, \frac{1}{2}, \frac{1}{4}\}$ is the downscale factors, and $D_s^k(\cdot)$ denotes the $k$-th layer features in $D_s$. 

\paragraph{Adversarial Loss.} It has been proved to be effective and critical in generating realistic textures in image restoration tasks. In this work, we use multi-scale discriminators and hinge loss as the objective function, defined as 
\begin{gather}
  \mathcal{L}_{GAN\_G} = \sum_s -\mathbb{E}(\hat{I}_H^s),\\
  \begin{split}
      \mathcal{L}_{GAN\_D} = \sum_s \Bigl\{\mathbb{E}\bigl[\max(0, 1 - D_s(I_H^s))\bigr] + \\ 
      \mathbb{E}\bigl[\max(0, 1 + D_s(\hat{I}_H^s))\bigr]\Bigr\},
  \end{split}
\end{gather}
For training stability, we incorporate spectral normalization \cite{miyato2018spectral} in both generator and discriminators.  
In summary, the final loss function for our generator network is \lxm{defined as}
\begin{equation}
 \mathcal{L}_G = \lambda_{ss} \mathcal{L}_{ss} + \lambda_{rec} \mathcal{L}_{rec} + \lambda_{adv} \mathcal{L}_{GAN\_G},
\end{equation}
where $\lambda$ are the weights for different terms. PSFR-GAN is trained by minimizing $\mathcal{L}_{G}$ and $\mathcal{L}_{GAN\_D}$ alternatively. 

\section{Implementation Details}
\subsection{Degradation Model}
According to previous works \cite{Li_2018_ECCV,xu2017learning} and common practice in SISR framework, we generate the LR image $I_L^r$ with the following degradation model:
\begin{equation}
  I_L^r = ((I_H \otimes \textbf{k}_\varrho)\downarrow_r + \textbf{n}_\delta)_{JPEG_q}, \label{equ:degrade} 
\end{equation}
where $\otimes$ represents the convolution operation between the HQ image $I_H$ and a blur kernel $\textbf{k}_\varrho$ with parameter $\varrho$. $\downarrow_r$ is the downsampling operation with a scale factor $r$. $\textbf{n}_\delta$ denotes the additive white Gaussian noise (AWGN) with a noise level $\delta$. $(\cdot)_{JPEG_q}$ indicates the JPEG compression operation with quality factor $q$. The hyper parameters $\varrho, r, \delta, q$ are randomly selected for each HR image $I_H$, and $I_L^r$ is generated online. More details are described in the appendix. After we get $I_L^r$, $I_L=(I_L^r)\uparrow_r$ is used for the parsing map prediction and face restoration. 

\begin{figure*}[htb]
    \centering
    \includegraphics[width=.99\linewidth]{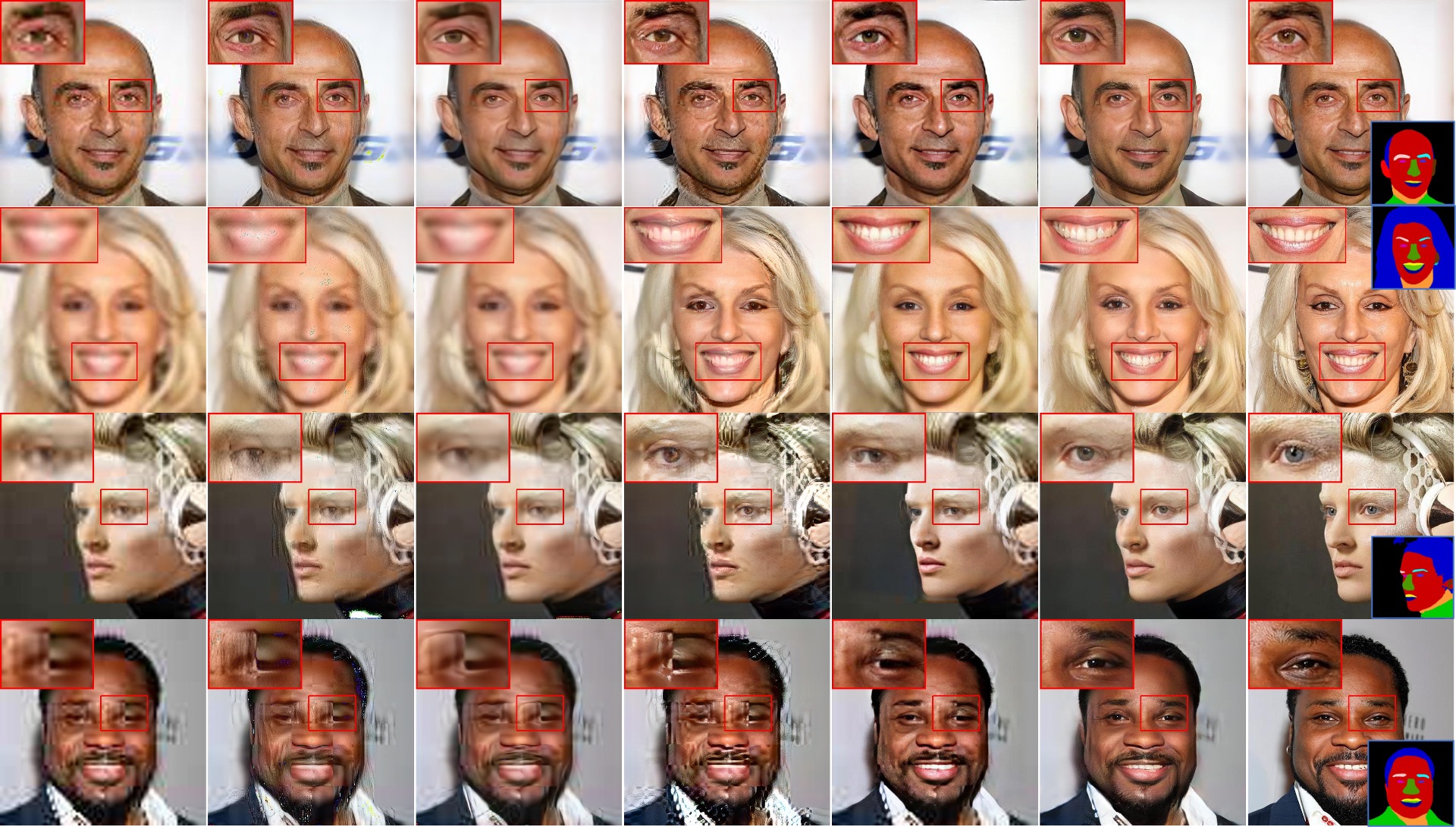}
    \begin{tabularx}{.99\linewidth}{*{7}{C}}
        LQ Input & ESRGAN & Super-FAN & HiFaceGAN & DFDNet & PSFR-GAN & GT
    \end{tabularx}
    \caption{Visual comparisons on CelebAHQ-Test dataset. The proposed PSFR-GAN shows the best results for LQ inputs with light degradation, severe degradation, large pose and different skin colors. The predicted parsing maps of \emph{LQ inputs} through pretrained FPN are overlapped at the corner of GT \lxm{images}. For better visualization experience, we only show results with top-5 FID scores in Table \ref{tab:quantitative}. Please zoom in to see the details.} \label{fig:comp-synth}
\end{figure*}

\subsection{Datasets}

\paragraph{Training Data.} We adopt the FFHQ \cite{karras2019style} as the training dataset. This dataset consists of $70,000$ high-quality images at a size of $1024\times1024$. All images are automatically cropped and aligned. We resize the images to $512\times512$ with bilinear downsampling as the ground-truth HR images, and synthesize the LQ inputs online with Eq. \ref{equ:degrade} where the parameters \lxm{are} randomly selected. 

\paragraph{Testing Data.} We construct two testing datasets, a synthetic one and a real one. For the synthetic test dataset, we randomly choose $2,800$ HQ images from CelebAHQ \cite{karras2017progressive} which has no identity intersection with FFHQ, and then generate the corresponding LQ images in the same way as training dataset. We denote this synthetic \lxm{test dataset} as \tb{CelebAHQ-Test}. For the real LQ test dataset, we collect $1,020$ faces smaller than $48\times48$ from CelebA \cite{liu2015faceattributes} and $106$ images provided by GFRNet \cite{Li_2018_ECCV}. The GFRNet-Test contains LR images from VGGFace2 \cite{Cao18} and IMDB-WIKI \cite{shrivastavaCVPR16ohem}. We also collect some old photos from the internet for testing. All images are cropped and aligned in the same manner as FFHQ, and then resized to $512\times512$ using bicubic upsampling. We merge all these images together and create a new dataset containing $1,157$ real LQ faces, denoted as \tb{PSFR-RealTest}. 

\subsection{Training Details}
We use Adam optimizer \cite{kingma2014adam} to train our networks. We choose $\beta_1=0.5, \beta_2=0.999$, and set the learning rate of the generator and discriminator to $0.0001$ and $0.0004$ respectively. The trade-off parameters of different losses are set as $\lambda_{rec}=10$, $\lambda_{ss}=100$ and $\lambda_{adv}=1$. The training batch size is set to $4$. All models were implemented by PyTorch \cite{pytorch} and trained on a Tesla V100 GPU. 

\section{Experiments}

\begin{table*}[htb]
\caption{Quantitative comparison on different restoration tasks with state-of-the-art methods. The test datasets are generated with FFHQ-Test using random parameters of each specific degradation type.} \label{tab:quantitative}
\centering
\begin{tabular}{c|c|ccc:cc|c}
\hline
\multirow{2}{*}{Task} & \multicolumn{1}{c|}{\multirow{2}{*}{Methods}} & \multicolumn{5}{c|}{CelebAHQ-Test} & \multicolumn{1}{c}{PSFR-RealTest} \\ \cline{3-8} 
& & PSNR$\uparrow$ & SSIM$\uparrow$ & MSSIM$\uparrow$ & LPIPS$\downarrow$ & FID$\downarrow$ & FID$\downarrow$ \\ \hline \hline
JPEG artifacts removal & ARCNN & 22.78 & 0.6538 & 0.7462 & 0.5862 & 133.38 & 124.46 \\ \hline
Deblur & DeblurGANv2 & 22.66 & 0.6587 & 0.7493 & 0.5546 & 113.85 & 97.42 \\ \hline
\multirow{3}{*}{Super-Resolution}
& ESRGAN & 21.95 & 0.6096 & 0.7293 & 0.5515 & 97.02 & 57.51 \\ 
& Super-FAN & 22.71 & 0.6527 & 0.7459 & 0.4908 & 94.95 & 65.45 \\ 
& WaveletSRNet & \underline{23.50} & \tb{0.6595} & 0.7542 & 0.5409 & 111.60 & 108.21 \\ \hline 
\multirow{3}{*}{Blind-Restoration}
& HiFaceGAN & 21.50 & 0.5495 & 0.6900 & 0.4569 & 57.81 & 56.48 \\ 
& DFDNet & 22.28 & \underline{0.6589} & \underline{0.7650} & \underline{0.3791} & \underline{37.34} & \underline{37.63} \\ 
& \tb{PSFR-GAN (ours)} & \tb{23.64} & 0.6557 & \tb{0.7740} & \tb{0.3042} & \tb{23.20} & \tb{30.39} \\ \hline 
\end{tabular}
\end{table*}

\begin{figure*}[htb]
  \centering
  \includegraphics[width=.99\linewidth]{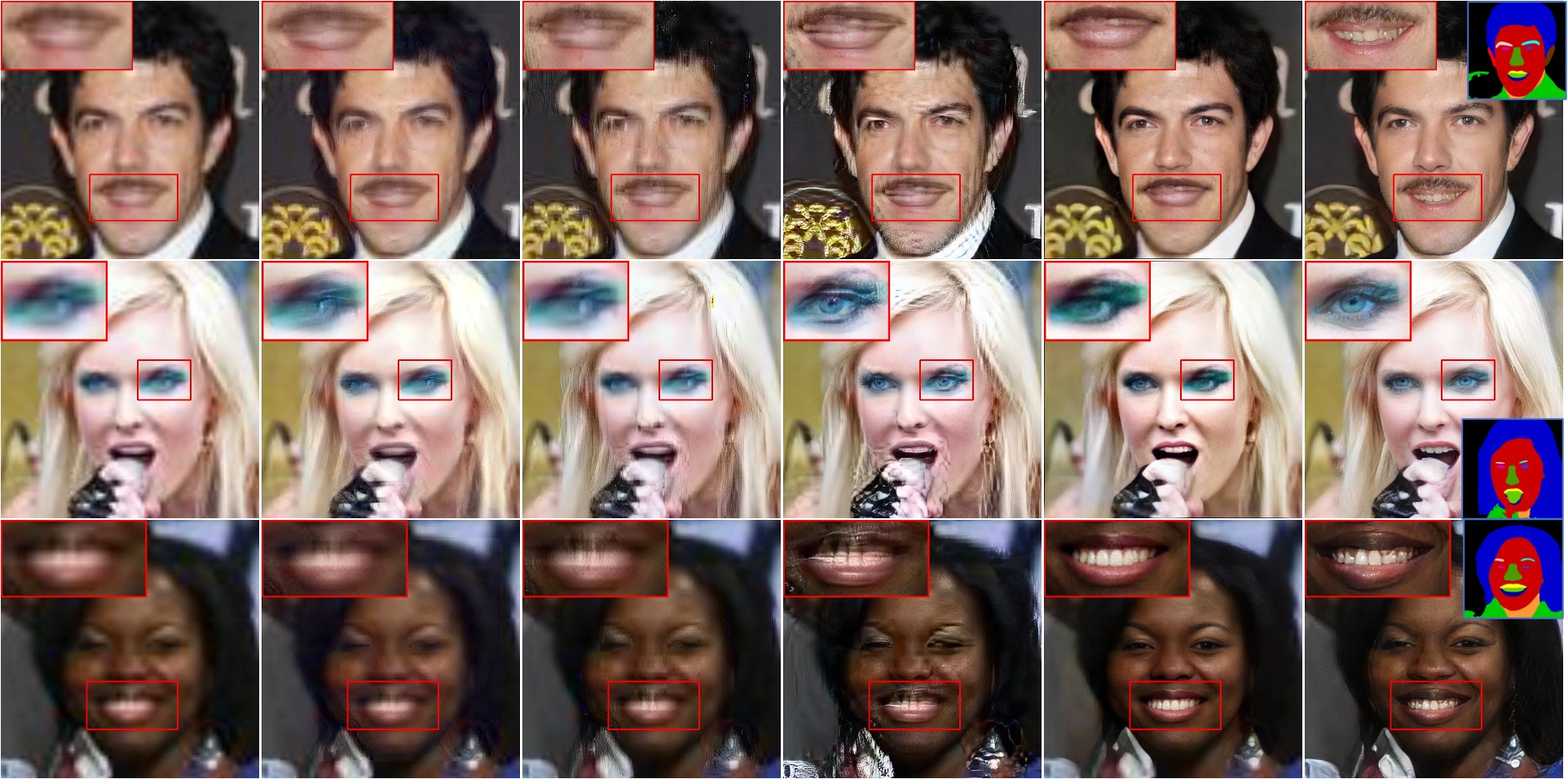}
  \begin{tabularx}{.99\linewidth}{*{6}{C}}
    LQ Input & ESRGAN & Super-FAN & HiFaceGAN & DFDNet & PSFR-GAN
  \end{tabularx}
  \caption{Visual comparisons on PSFR-RealTest dataset. Results of the proposed PSFR-GAN are clearer and more realistic. We show results with top-5 FID scores in Table \ref{tab:quantitative}. More results are provided in appendix.}
  \label{fig:comp-real}
\end{figure*}

In this section, we conduct experiments to compare our framework with other methods on both synthetic and real LQ test datasets, and carry out extensive ablation studies to evaluate the effectiveness of the multi-scale parsing map guidance and {\em semantic aware style loss}.  

\subsection{Evaluation Metrics}
For CelebAHQ-Test with ground truth, we take the widely used PSNR, SSIM and MSSIM metrics. However, these pixel space metrics prefer smooth results and are not consistent with human perception, therefore we also adopt LPIPS score \cite{zhang2018perceptual} to evaluate the perceptual realism of generated faces. For natural LQ images without ground truth, we use FID score \cite{heusel2017gans} to measure the statistic distance between the restoration results and a reference HQ face dataset. Compared with other no reference metrics such as NIQE \cite{niqe-score-2012} and IS (Inception Score) \cite{salimans2016improved} which focus on natural images, FID utilizes a reference HQ face dataset and gives better measurement. We use the ground truth images from CelebAHQ-Test as the reference dataset to evaluate results of PSFR-RealTest. We also provide FID scores for results of CelebAHQ-Test for reference. 

\subsection{Comparison on Synthetic Datasets} 

We first evaluate the performance of different methods on the synthetic CelebAHQ-Test dataset. Following HiFaceGAN \cite{Yang2020HiFaceGANFR}, we compare the proposed PSFR-GAN with architectures designed for different restoration tasks: ARCNN \cite{arcnn} for JPEG artifacts removal; DeblurGANv2 \cite{deblurganv2} for image deblurring; ESRGAN \cite{wang2018esrgan} for natural image super-resolution; Super-FAN \cite{bulat2017super} and WaveletSRNet \cite{huang2017wavelet} for face super-resolution; and recent methods HiFaceGAN \cite{Yang2020HiFaceGANFR} and DFDNet \cite{Li_2020_ECCV} for blind face restoration. Table \ref{tab:quantitative} shows the performance of both statistical metrics (PSNR, SSIM, MSSIM) and perceptual metrics (LPIPS, FID). We can observe that methods designed for specific tasks generally show low perceptual scores than blind face restoration methods. Both DFDNet and PSFR-GAN utilize extra face prior, and their performance is much better than HiFaceGAN. Compared with DFDNet, the proposed PSFR-GAN outperforms by a large margin in terms of most evaluation metrics, especially in FID score ($50\%$ improvement), which indicates the superiority of \lxm{our proposed} PSFR-GAN. 

The visualization examples in Fig. \ref{fig:comp-synth} help us understand the quantitative results better. We show three kinds of LQ inputs in each row of Fig. \ref{fig:comp-synth}: LQ with light degradation, LQ with severe degradation, LQ with large pose and different skin color. It can be observed that results of both ESRGAN and Super-FAN are over smoothed and fail to recover clear face components and textures compared with blind face restoration methods, see $2$-nd and $3$-rd rows. This indicates that methods designed for specific task cannot handle LQ inputs with complicated blind degradation. The results of HiFaceGAN are clearer and sharper but contain too much artifacts in the mouth, eyes and backgrounds. This is most likely because of the unstable training of GAN without face prior guidance. Different from PSFR-GAN which utilizes parsing map, DFDNet needs to first detect the LQ face components and then matches them to a HQ dictionary. It may fail to find the correct reference when the LQ are too blurry or with large pose. For example, the results of DFDNet in the $3$-rd row seems to have two eyeballs which do not exist in the other methods. Meanwhile, PSFR-GAN incorporates the semantic information through parsing map which is more generic than explicit HQ reference. Therefore, the results of PSFR-GAN are more realistic and robust with all kinds of LQ inputs. Moreover, our results are generated using the predicted parsing maps of the pretrained FPN, see last column in Fig. \ref{fig:comp-synth}. This means PSFR-GAN does not need extra information such as HQ dictionaries during the test time, making PSFR-GAN available in most situations.

\subsection{Comparison on Real World LQ Images}

The final target of all methods is to restore real world LQ face images. To evaluate the generalization ability of different methods, we also compare the performance of PSFR-GAN on PSFR-RealTest dataset with methods in Table \ref{tab:quantitative}. The same as previous results on CelebAHQ-Test, methods designed for blind face restoration still outperform the others. FID score of PSFR-GAN surpasses other methods by a large margin, and is $20\%$ higher than the second best result of DFDNet. Fig. \ref{fig:comp-real} gives some examples from PSFR-RealTest dataset. We can observe that there are many artifacts in the results of HiFaceGAN. DFDNet seems to have difficulties finding the correct component references for real LQ inputs, for example, teeth in the first row and right eye in the second row. In contrast, the results of our method is much more natural with few conspicuous artifacts. This is because PSFR-GAN generates the results in a progressive way, and the multi-scale parsing maps provide multi-scale style guidance which makes PSFR-GAN able to synthesis realistic textures for each semantic regions, \eg, teeth textures and eyes. The parsing map results in the last column illustrate that the pretrained FPN works well for real LQ inputs. We show more results of PSFR-GAN on PSFR-RealTest datasets in appendix, which demonstrate that PSFR-GAN is practical for real world scenarios. 

\begin{figure}[t]
  \includegraphics[width=.99\linewidth]{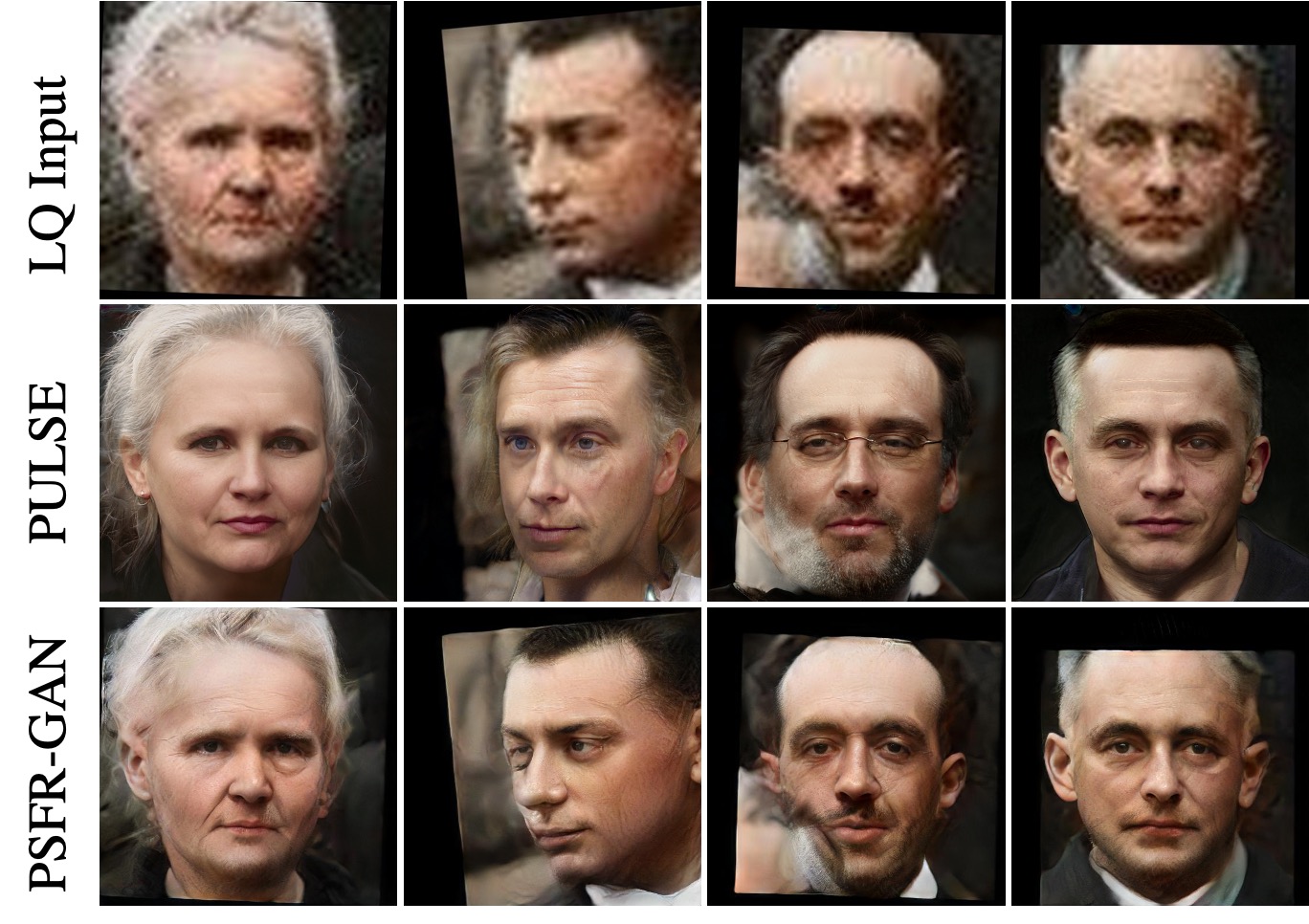}
  \caption{Visual comparison between PULSE and PSFR-GAN. Complete results are provided in the appendix. Please zoom in to see the details.}
  \label{fig:comp-pulse}
\end{figure}

\subsection{Comparison with PULSE} 

PULSE \cite{menon2020pulse} is a recent popular method for face restoration. Different from other methods, PULSE is an optimization based method which needs carefully finetune for each LQ input. Therefore, it is unfair to compare the quantitative result of PULSE with others on PSFR-RealTest. Instead, we follow HiFaceGAN and use the historic group photograph of famous physicists taken at the 5th Solvay Conference 1927 for visual comparison. We carefully finetune PULSE on these photos and get the best results as much as we can. Even so, we still observe several typical failure cases of PULSE shown in Fig. \ref{fig:comp-pulse}: (1) age mismatch and shape deformation in the first column; (2) large pose in the second column; (3) background interference in the third column. The final column shows the best result of the test photos, but there are many slight shape changes in the eyes, nose and mouth which make it look like another person. To conclude, although PULSE can generate better details, for example hair textures, the results of our PSFR-GAN \lxm{are} still much better than PULSE in face restoration. Besides, PSFR-GAN is 40 times faster than PULSE (0.1s vs. 4s) to process the same image on GPU. Complete results of Solvay conference are in the appendix.

\section{Discussions and Ablation Study} 

\begin{figure}[t]
    \begin{subfigure}{.99\linewidth}
        \centering
        \includegraphics[width=.99\linewidth]{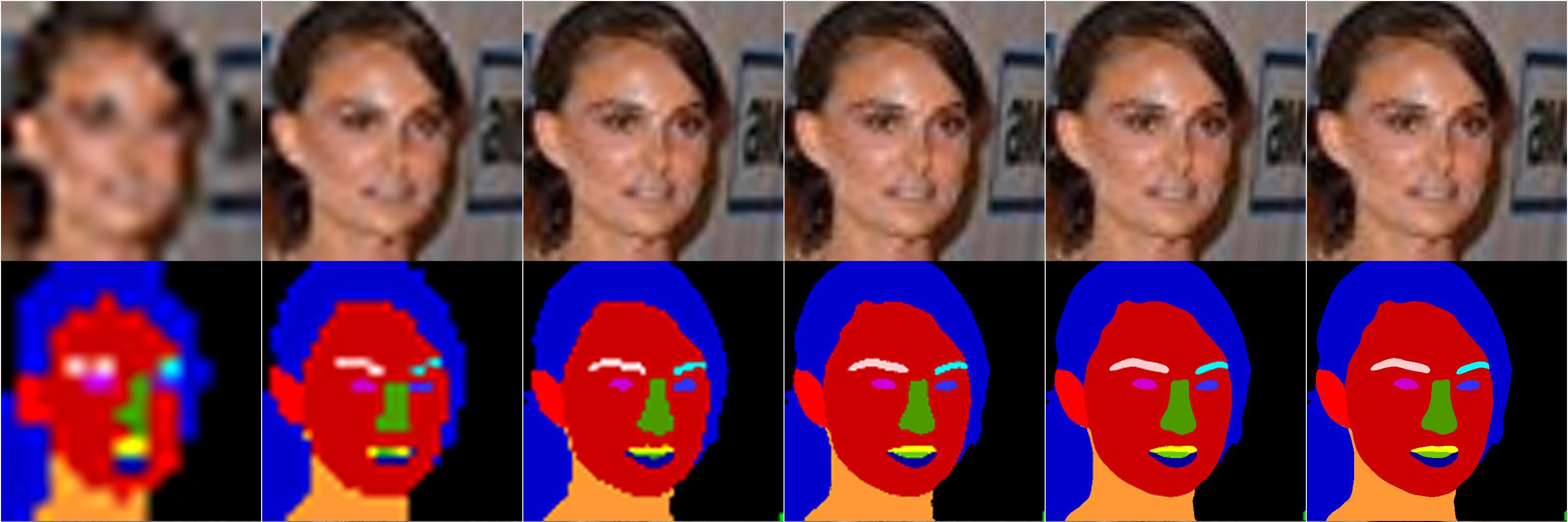}
        \begin{tabularx}{.99\linewidth}{*{6}{C}}
            16 & 32 & 64 & 128 & 256 & 512 
        \end{tabularx}
        \caption{Multi-scale LQ and parsing map input pairs $(I_L^i, I_P^i)$. } \label{fig:psfr-analysis-0}
    \end{subfigure}
    \begin{subfigure}{.99\linewidth}
        \centering
        \includegraphics[width=.99\linewidth]{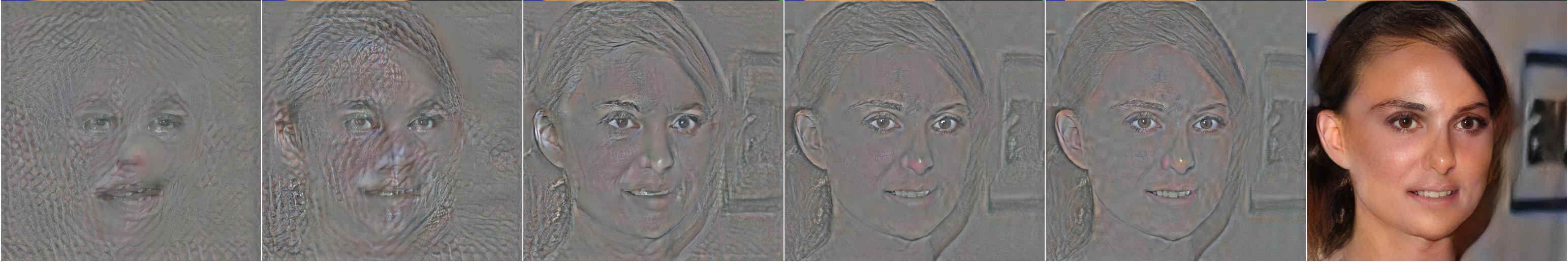}
        \scriptsize
        \begin{tabularx}{.99\linewidth}{*{6}{C}}
            $(I_L^1, I_P^1)$ & $+(I_L^2, I_P^2)$ & $+(I_L^3, I_P^3)$ & $+(I_L^4, I_P^4)$ & $+(I_L^5, I_P^5)$ & $+(I_L^6, I_P^6)$ 
        \end{tabularx}
        \caption{Results of adding multi-scale inputs $(I_L^i, I_P^i)$ progressively. } \label{fig:psfr-analysis-a}
    \end{subfigure}
    \begin{subfigure}{.99\linewidth}
        \centering
        \includegraphics[width=.99\linewidth]{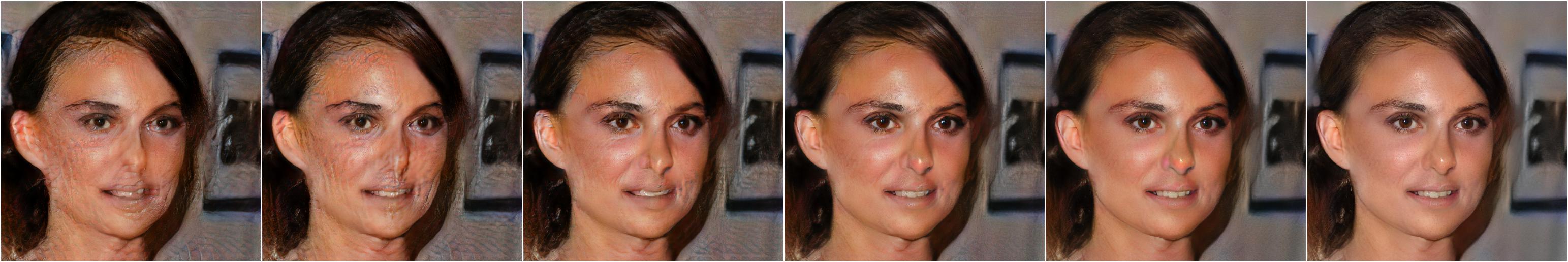}
        \caption{Keep all scales of LQ inputs $I_L^i$, and add different scales of parsing maps $I_P^i$ progressively. } \label{fig:psfr-analysis-b}
    \end{subfigure}
    \begin{subfigure}{.99\linewidth}
        \centering
        \includegraphics[width=.99\linewidth]{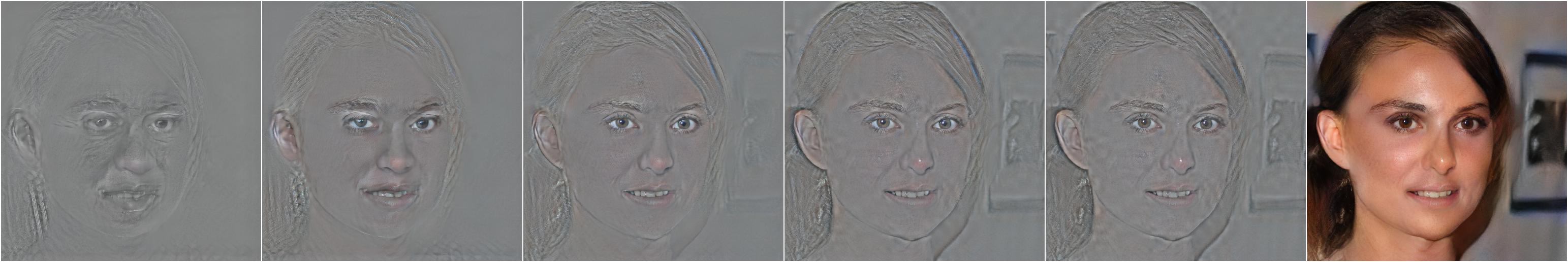}
        \caption{Keep all scales of parsing map inputs $I_P^i$, and add different scales of LQ inputs $I_L^i$ progressively. } \label{fig:psfr-analysis-c}
    \end{subfigure}
    \caption{Analysis of PSFR-GAN. Zoom in to see details.} \label{fig:psfr-analysis}
\end{figure}

\subsection{Analysis of PSFR-GAN} 

\paragraph{Why PSFR-GAN works ?} There are two key designs in PSFR-GAN: (1) the features are modulated progressively from coarse-to-fine; (2) the multi-scale input pairs $I_L$ and $I_P$ \lxm{work} together to provide sufficient color, shape and semantic information. In Fig. \ref{fig:psfr-analysis-a}, we show how details are progressively added to the final outputs by first zeroing all inputs and then adding them back from coarse to fine. We can observe that the restoration process is consistent with our hypothesis, and happens in the following order: high level semantic information comes first, then the mid-level shape and edges, and finally low-level color and details. Similarly, we analyze the effect of $I_P^i$ and $I_L^i$ separately in Fig. \ref{fig:psfr-analysis-b} and Fig. \ref{fig:psfr-analysis-c}. The first column of Fig. \ref{fig:psfr-analysis-b} and Fig. \ref{fig:psfr-analysis-c} show the result of using $I_L^i$ and $I_P^i$ as inputs separately. We can see that in the first image of Fig. \ref{fig:psfr-analysis-b}, the nose and mouth borders are not clear and there are many artifacts in the cheek region. This indicates that network without parsing map as inputs only makes the bicubic results (first row and last column in Fig. \ref{fig:psfr-analysis-a}) sharper and has difficulties in understanding the semantic meaning of each region. When we add $I_P^i$ progressively, the artifacts are gradually removed and the edges are clearer. As for Fig. \ref{fig:psfr-analysis-c}, we can observe that the semantic regions are clear in all stages, for example the nose and teeth part. Then, color and texture details are added with the entry of $I_L^i$. In summary, PSFR-GAN restores the LQ images in a progressive way by modulating the features with multi-scale inputs, where $I_L^i$ provides the low level color and texture informaiton and $I_P^i$ contributes the semantic and shape information. 

\begin{figure}[t]
    \centering
    \includegraphics[width=.99\linewidth]{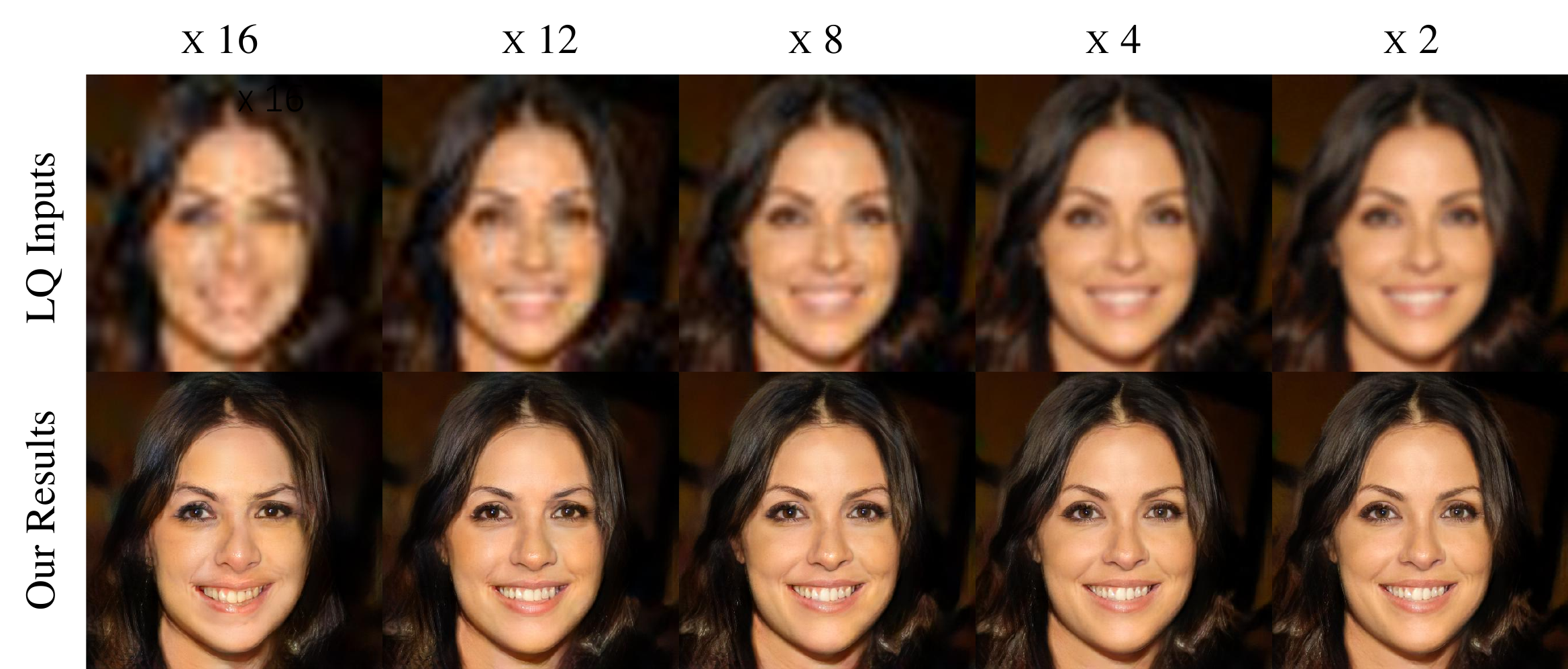}
    \caption{Results for different upscale factors while fixing other degradation parameters.}
    \label{fig:diff_degrade}
\end{figure}

\paragraph{Robustness to different degrees of degradation} As an example, we verify the effectiveness of the PSFR-GAN for different upscale factors in Eq. \ref{equ:degrade} and fix other parameters. We can observe from Fig. \ref{fig:diff_degrade} that our network works well for upscale factors $\leq 12$ and can produce reasonable result for $\times16$, which demonstrates the robustness of PSFR-GAN. 

\subsection{Ablation Study}

\begin{figure}[t]
  \includegraphics[width=.99\linewidth]{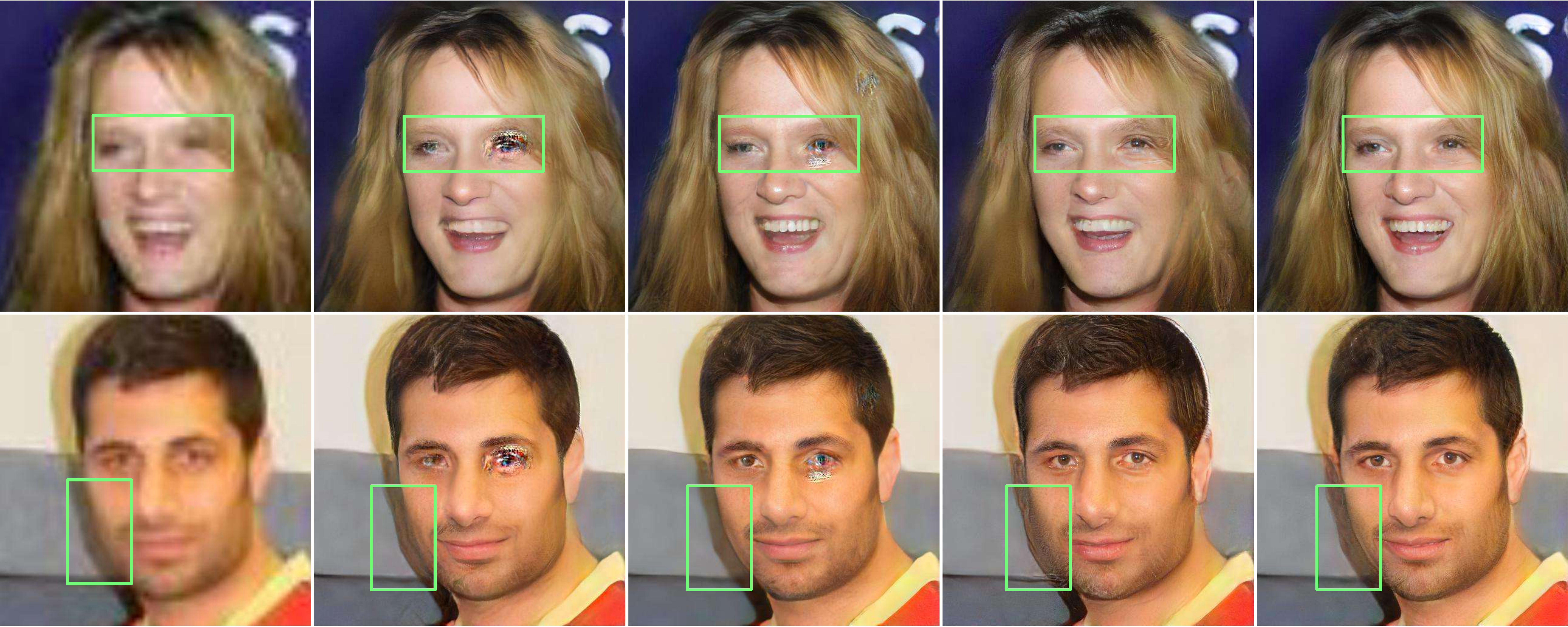}
  \begin{tabularx}{.99\linewidth}{*{5}{C}}
    LQ & Baseline & + $I_P$ & + $\mathcal{L}_{ss}$ & \scriptsize{+ $I_P$ + $\mathcal{L}_{ss}$} 
  \end{tabularx}
  \caption{Visual comparison between different variations of our model. Please zoom in to see the details.}
  \label{fig:exp-ablation}
\end{figure}

To explore the effectiveness of parsing map guidance and {\em semantic-aware style loss} $\mathcal{L}_{ss}$, we evaluate four variants of our framework in Table \ref{tab:ablation}: \bsf{A}, baseline model with only $I_L$ as inputs; \bsf{B}, baseline model with $(I_L, I_P)$ as inputs but without $\mathcal{L}_{ss}$; \bsf{C}, baseline model with $I_L$ as inputs and $\mathcal{L}_{ss}$; \bsf{D}, the proposed PSFR-GAN. \lxm{We can observe that parsing map plays an important role in face image restoration and achieves the most improvements, and the $\mathcal{L}_{ss}$ can also benefit the restoration results. With the combination of them, our PSFR-GAN can achieve the best performance.}
Figure \ref{fig:exp-ablation} shows some example results on PSFR-RealTest. We can observe from the bottom row that models without parsing map (\ie, model \bsf{A} and \bsf{C}) fail to generate clear shapes when the face border in LQ image is not clear. On the other hand, models without $\mathcal{L}_{ss}$ (\ie, model \bsf{A} and \bsf{B}) produce apparent artifacts especially in the eyes. In contrast, model \bsf{D} with parsing map and $\mathcal{L}_{ss}$ has none of the above flaws. It can be inferred from the above observation that 1) parsing map helps to regularize face structure, and 2) $\mathcal{L}_{ss}$ helps to synthesize realistic textures for each semantic region. 

\begin{table}[t]
\centering
\caption{Ablation study of the proposed method.} \label{tab:ablation}
\begin{tabularx}{.8\linewidth}{c|l|C}
\hline
ID & Model Variations & FID$\downarrow$ \\ \hline \hline 
\bsf{A} & Baseline with $I_L$ & 47.48 \\ \hline
\bsf{B} & + $I_P$ & 32.37  \\ \hline
\bsf{C} & + $\mathcal{L}_{ss}$ & 34.22 \\ \hline
\bsf{D} & + $I_P$ + $\mathcal{L}_{ss}$ & \textbf{30.39} \\ \hline
\end{tabularx}
\end{table}

\section{Conclusion}

This paper proposes a multi-scale progressive face restoration network, named PSFR-GAN, which restores LQ face inputs in a coarse to fine manner through semantic-aware style transformation. We also proposed the {\em semantic-aware style loss} based on original gram matrix loss. Experiments on both synthetic and real LQ test datasets demonstrate the superority and robustness of our PSFR-GAN. By pretraining the face parsing network (FPN) for LQ inputs, our framework can generate high-resolution and realistic HQ outputs without requiring extra inputs. In summary, PSFR-GAN provides a robust and easy-to-use solution for face restoration in real-world scenarios. 

\section{Acknowledgements}
This work was partially supported by Alibaba DAMO Academy, Hong Kong RGC RIF grant (R5001-18), and Hong Kong RGC GRF grant (project\# 17203119).

{\small
\bibliographystyle{ieee_fullname}
\bibliography{egbib}
}

\newpage
\appendix
\section*{\Large\appendixname}

\section{Pretrain of LQ Face Parsing Network} 
\subsection{Network Architecture}
\begin{figure*}[ht]
  \includegraphics[width=.99\linewidth]{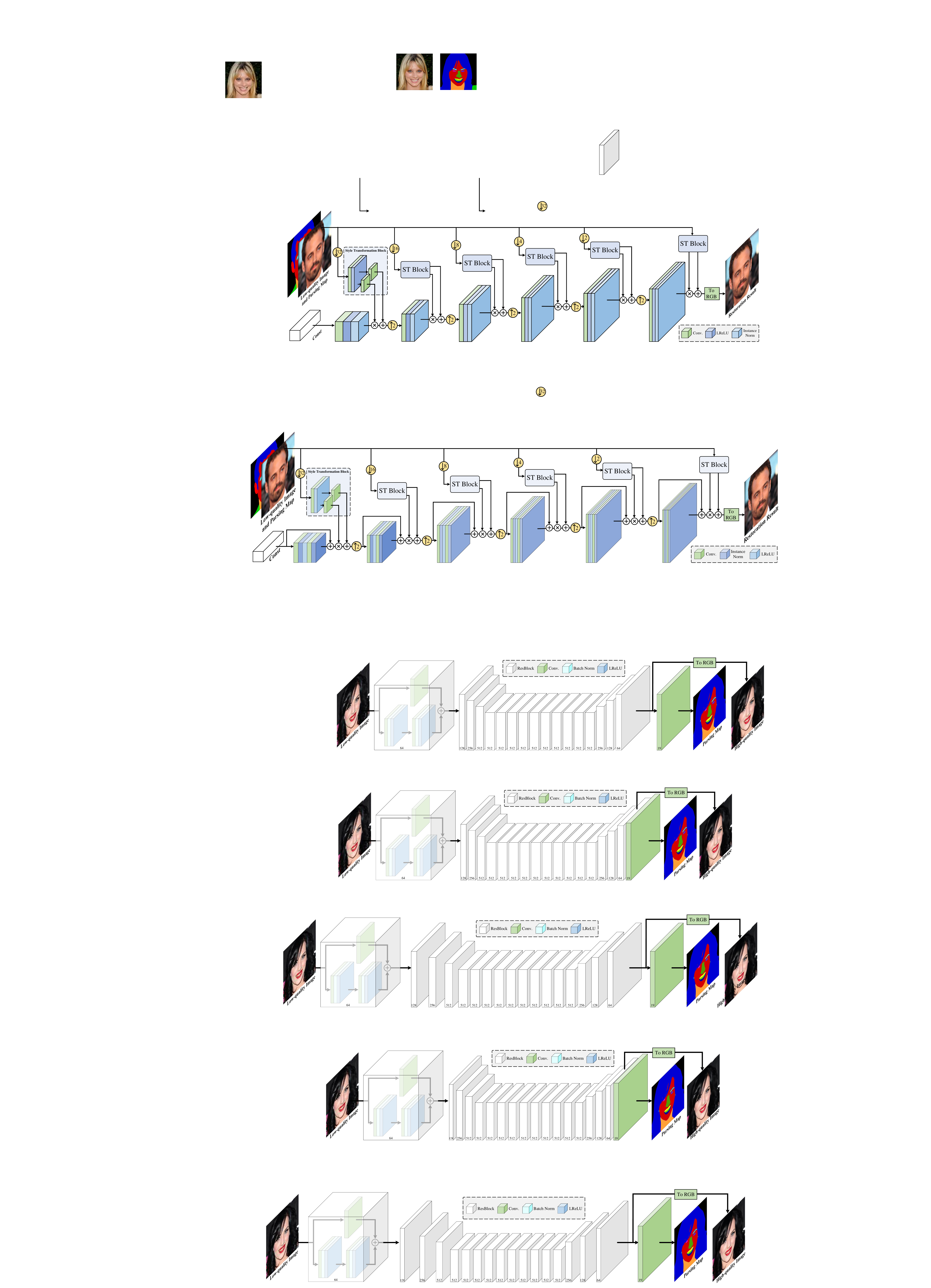}
  \caption{Architecture details of face parsing network (FPN).}
  \label{fig:method-parse_arch}
\end{figure*}

Given a LQ face image of any size, we first upsample it to $512\times512$ and treat it as the input $I_L$. FPN is trained to produce a parsing map $\hat{I}_P$ and a HR face $\hat{I_{H}}$ that approximate the ground-truth parsing map $I_P$ and ground-truth HR face $I_H$ respectively, \ie,
\begin{eqnarray}
  \theta_p&=&\argmin_{\theta_p} \mathcal{L}_{parse}(\hat{I}_P, I_P) + \mathcal{L}_{pix}(\hat{I_{H}}, I_H), \label{equ:loss_parse}
\end{eqnarray}
where $\theta_p$ denotes the parameters of FPN, $\mathcal{L}_{parse}$ is the parsing loss, and $\mathcal{L}_{pix}$ is the pixel space L2 loss. As shown in Fig. \ref{fig:method-parse_arch}, FPN adopts an encoder-resnet-decoder architecture. It begins with 4 downsample blocks, followed by 10 resnet blocks and 4 upsample blocks. Finally, two output convolution layers are used to generate $\hat{I}_P$ and $\hat{I_H}$. We adopt BatchNorm-LeakyRelu after every convolution layer. 

\begin{figure*}[ht]
  \includegraphics[width=.99\linewidth]{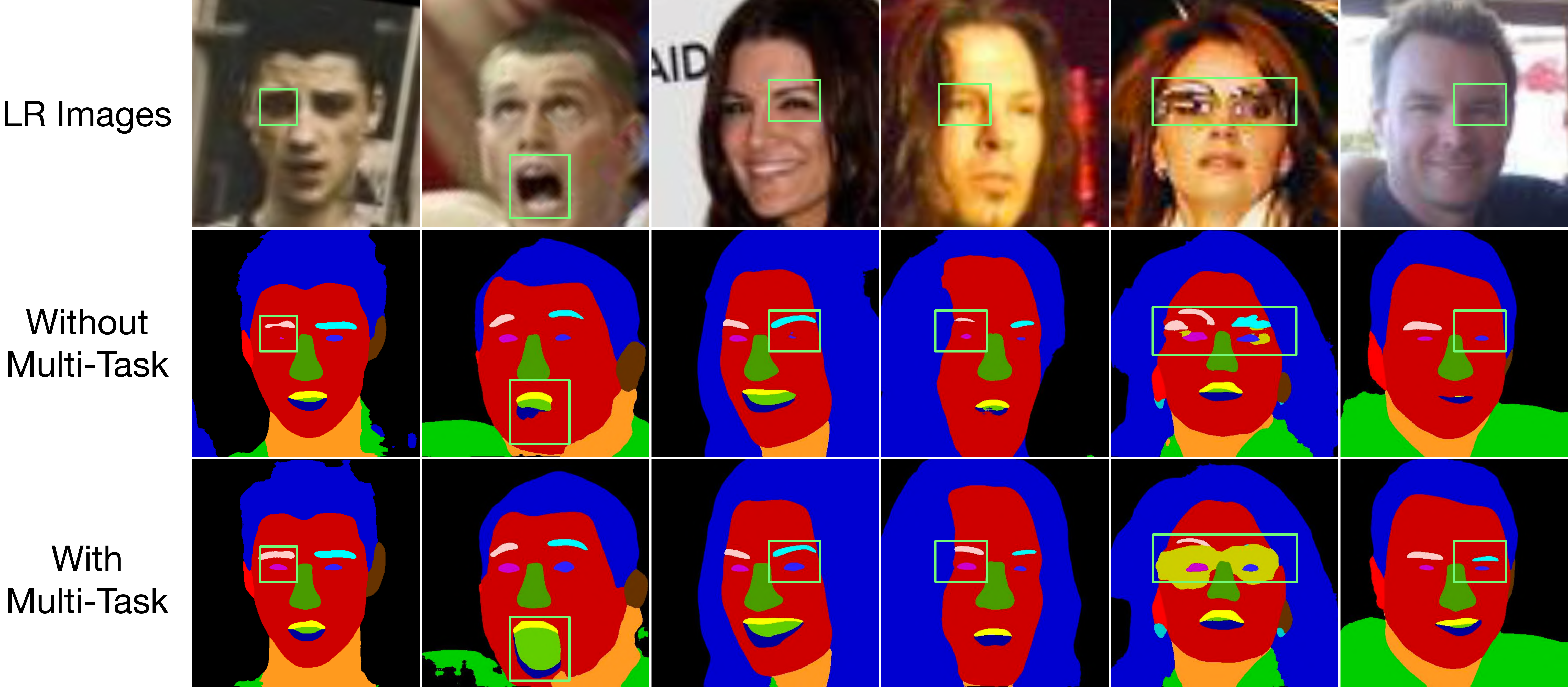}
  \caption{Comparison of parsing results of natural LR faces with and without supervision of $I_H$ .}
  \label{fig:method-parse_eg}
\end{figure*}

We use multi-task learning for FPN because we found that $\mathcal{L}_{pix}$ is quite helpful for the prediction of $\hat{I_P}$. Since $I_L$ is degraded, both the pixel values and pattern of the face components are not clear and stable. The network is not able to understand the meaning of each label without the extra supervision of $I_H$. Fig.~\ref{fig:method-parse_eg} shows that the parsing results with multi-task learning are much better than that without it, especially in the eyes and eyebrows.  

\subsection{Datasets and Implementation}
We use CelebA-Mask-HQ \cite{CelebAMask-HQ} to train FPN. The CelebA-Mask-HQ contains $30,000$ HR faces with a size of $1024\times1024$ selected from the CelebA \cite{liu2015faceattributes} dataset. Each image has a segmentation mask of facial attributes corresponding to CelebA. The masks of CelebA-Mask-HQ are manually-annotated with a size of $512 \times 512$ and 19 classes including background, skin, nose, eyes (left and right), eyebrows (left and right), ears (left and right), mouth, lips (up and bottom), hair, hat, eyeglass, earring, necklace, neck, and cloth. The whole dataset is split into a training set ($24,183$ images), a validation set ($2,993$ images), and a test set ($2,824$ images). We use the training set as ground-truth HQ faces and parsing maps, and the LQ faces are generated online with Eq.~\ref{equ:degrade}. 

We use Adam optimizer \cite{kingma2014adam} to train the FPN. We set $\beta_1=0.9, \beta_2=0.999$ and learning rate to $0.0002$. The training batch size is set to $8$. 

\section{Degradation Model} \label{sec:degradation}

As described in the paper, our degradation model used the following equation:
\begin{equation}
  I_L^r = ((I_H \otimes \textbf{k}_\varrho)\downarrow_r + \textbf{n}_\delta)_{JPEG_q}, \label{equ:degrade} 
\end{equation}
where 
\begin{itemize}
\item $\textbf{k}_\varrho$ is the blur kernel. We randomly choose one of the following four kernels: Gaussian Blur ($3 <= \varrho <= 15$),  Average Blur ($3 <= \varrho <= 15$), Median Blur ($3 <= \varrho <= 15$), Motion Blur ($5 <= \varrho <= 25$);
\item $\downarrow_s$ is the downsample operation. The scale factor $r$ is randomly selected in $[\frac{32}{512}, \frac{256}{512}]$; 
\item $\textbf{n}_\delta$ is the addictive white gaussian noise (AWGN) with $0 <= \delta <= 0.1\times255$;
\item $JPEG_q$ is the JPEG operation. The compression level is randomly chosen from $[10, 65]$, in which higher means stronger compression and lower image quality.  
\end{itemize}

We implement the degradation model using \texttt{imgaug}\footnote{\url{https://github.com/aleju/imgaug}} library with code snippets in Fig.~\ref{fig:degradation-code}.

\begin{figure*}[ht]
  \begin{python}
    import imgaug as ia
    import imgaug.augmenters as iaa
    scale_size = random(32, 256)
    org_size = 512
    aug_seq = iaa.Sequential([
            iaa.Sometimes(0.5, iaa.OneOf([
                iaa.GaussianBlur((3, 15)), iaa.AverageBlur(k=(3, 15)),
                iaa.MedianBlur(k=(3, 15)), iaa.MotionBlur((5, 25))
            ])),
            iaa.Resize(scale_size, interpolation=ia.ALL),
            iaa.Sometimes(0.2, iaa.AdditiveGaussianNoise(loc=0, scale=(0.0, 0.1*255), per_channel=0.5)),
            iaa.Sometimes(0.7, iaa.JpegCompression(compression=(10, 65))),
            iaa.Resize(org_size),
        ])
  \end{python}
  \caption{Code snippets for degradation model.} \label{fig:degradation-code}
\end{figure*}

\section{More Results}

In this section, we show more results on PSFR-RealTest and Solvay conference test. We mainly compare our model with DFDNet because they provide public codes and test models, and their results are current state-of-the-art. We also provide carefully finetuned results of PULSE on Solvay test. 

\subsection{Results of PSFR-RealTest}

\begin{figure*}[h]
  \centering
  \includegraphics[width=.99\textwidth]{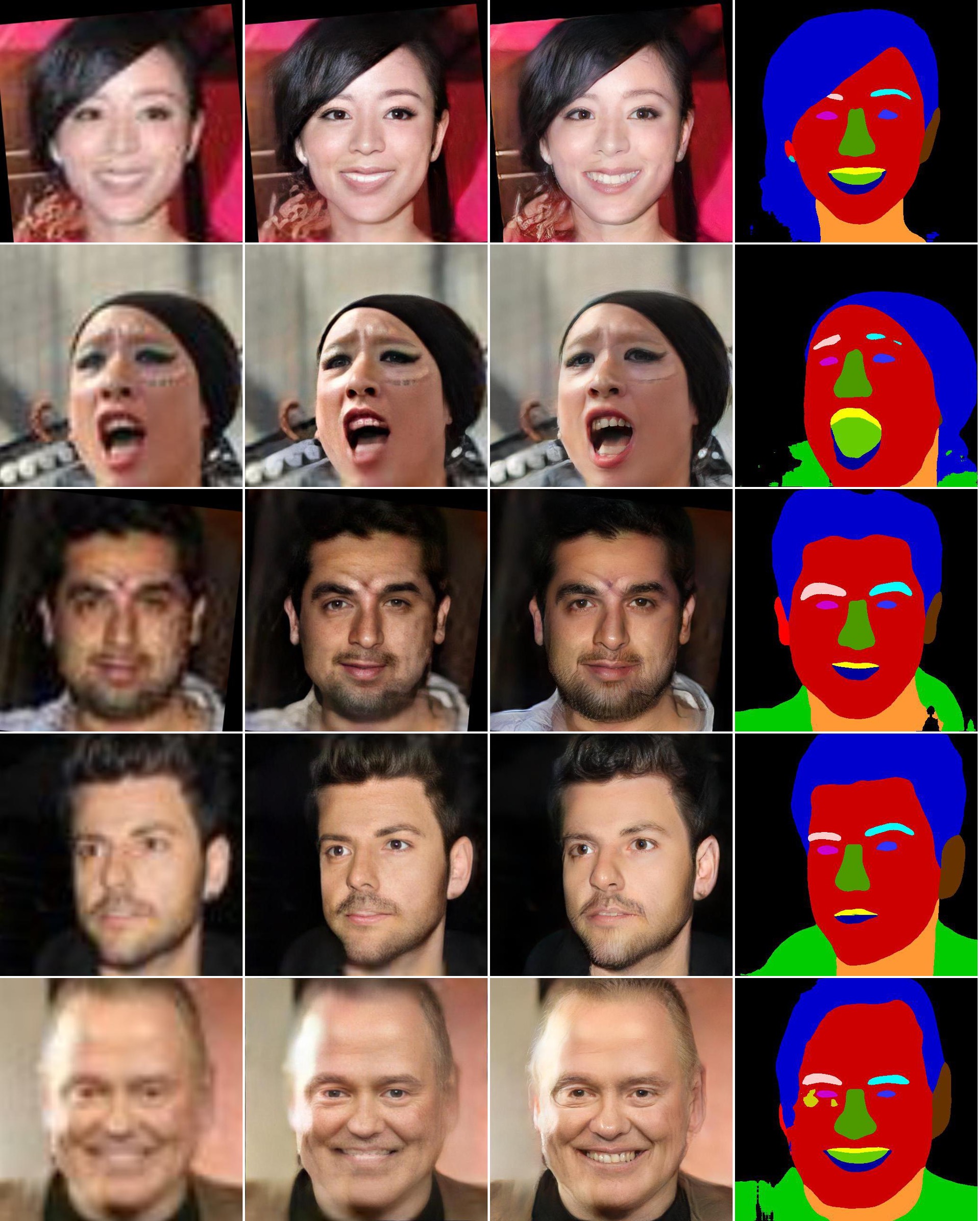}
  \begin{tabularx}{.99\linewidth}{*{4}{C}}
    LQ & DFDNet & PSFR-GAN (ours) & Parsing Map (ours)
    \end{tabularx}
  \caption{More results from PSFR-RealTest Dataset.}
  \label{fig:psfr-realtest-eg1}
\end{figure*}

\begin{figure*}[h]
  \centering
  \includegraphics[width=.99\textwidth]{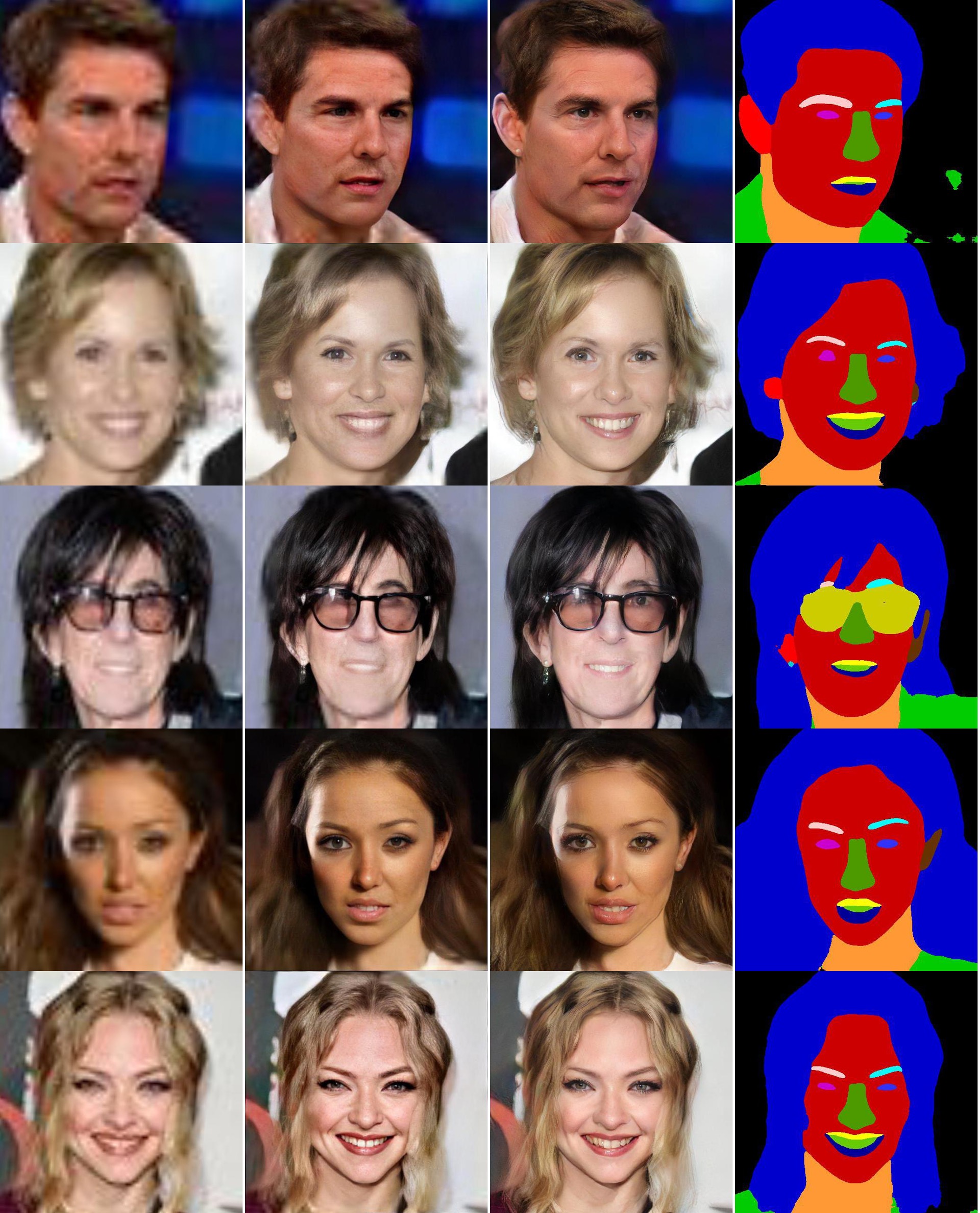}
  \begin{tabularx}{.99\linewidth}{*{4}{C}}
    LQ & DFDNet & PSFR-GAN (ours) & Parsing Map (ours)
    \end{tabularx}
  \caption{More results from PSFR-RealTest Dataset.}
  \label{fig:psfr-realtest-eg2}
\end{figure*}

\begin{figure*}[h]
  \centering
  \includegraphics[width=.99\textwidth]{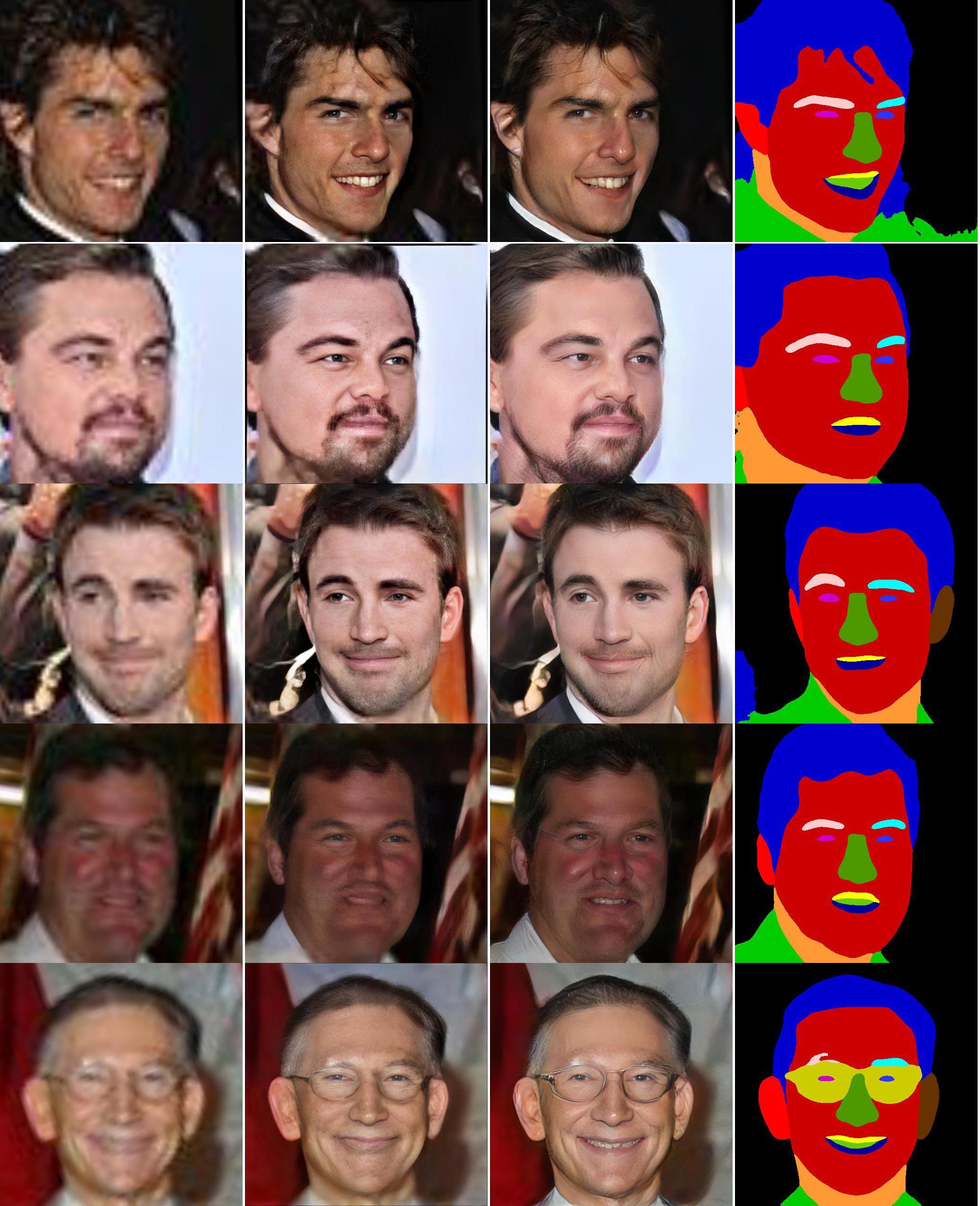}
  \begin{tabularx}{.99\linewidth}{*{4}{C}}
    LQ & DFDNet & PSFR-GAN (ours) & Parsing Map (ours)
    \end{tabularx}
  \caption{More results from PSFR-RealTest Dataset.}
  \label{fig:psfr-realtest-eg3}
\end{figure*}

Fig. \ref{fig:psfr-realtest-eg1}, Fig. \ref{fig:psfr-realtest-eg2} and Fig. \ref{fig:psfr-realtest-eg3} show more examples from PSFR-RealTest dataset. 

\subsection{Results of Solvay Conference Test}

\begin{figure*}[h]
  \centering
  \advance\leftskip-1cm
  \includegraphics[width=1.12\textwidth]{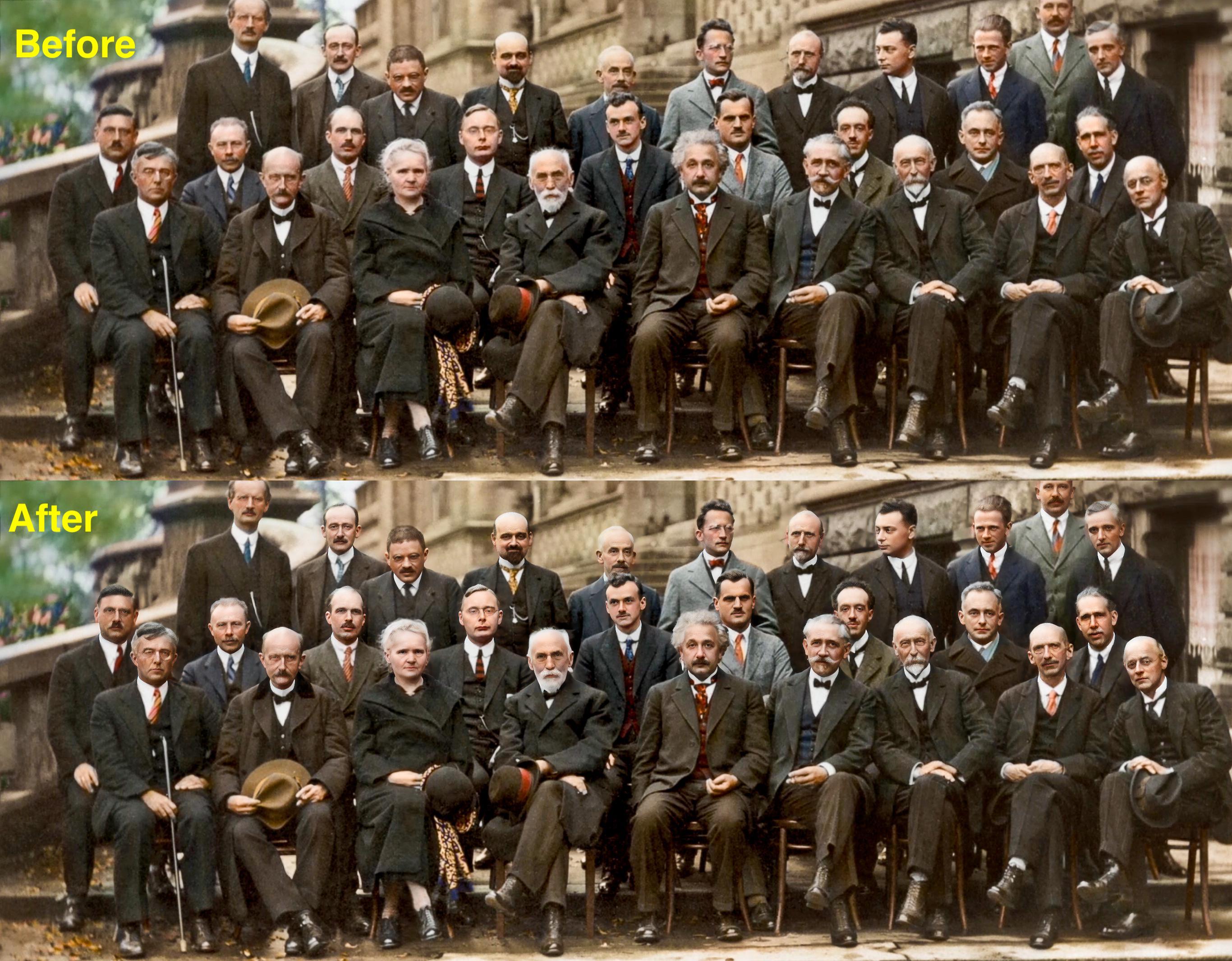}
  \caption{Overall result of the $5$-th Solvay conference taken in 1927. Please zoom in to see the details.}
  \label{fig:solvay-test-overall}
\end{figure*}

\begin{figure*}[h]
  \centering
  \includegraphics[width=.99\textwidth]{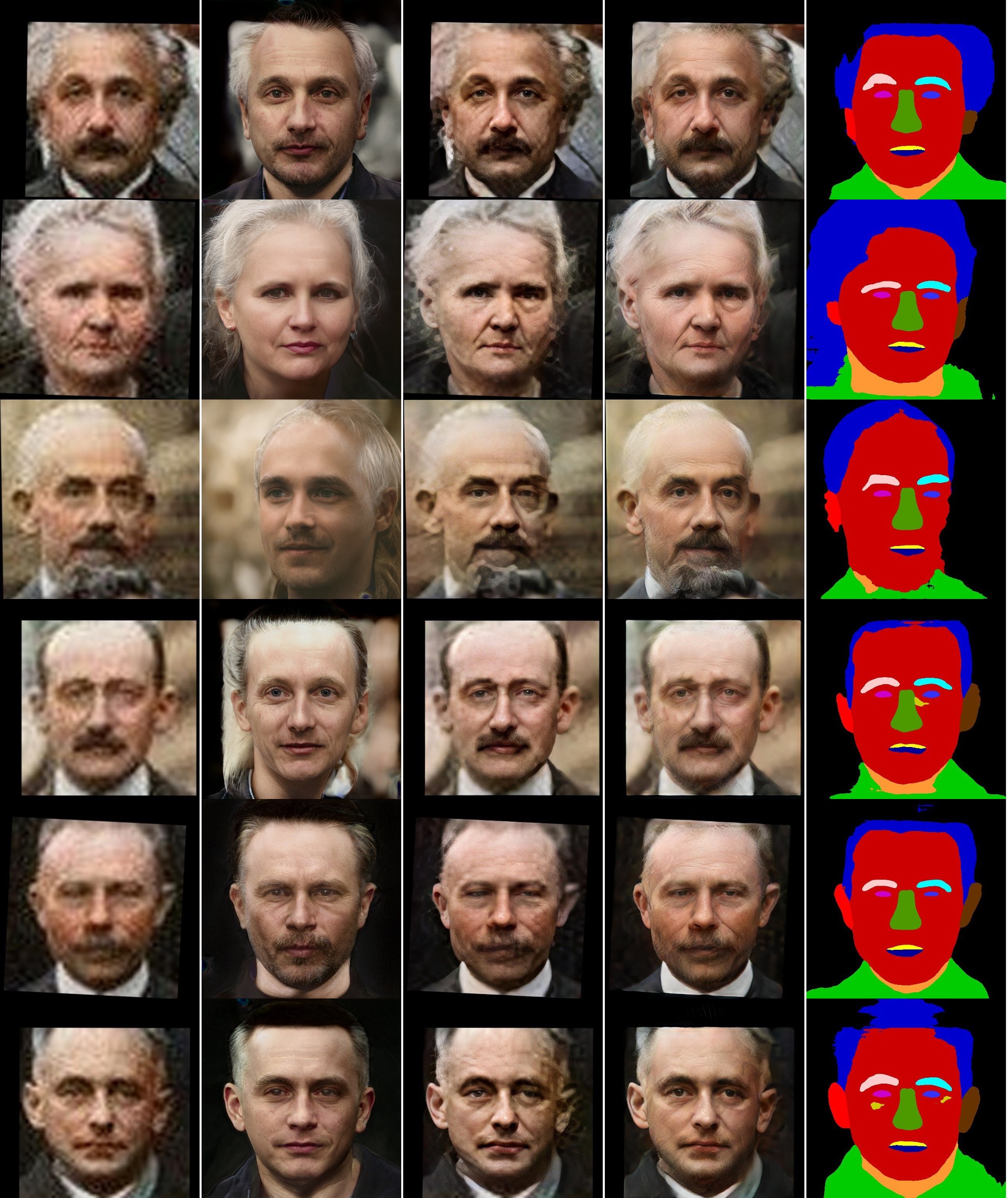}
  \begin{tabularx}{.99\linewidth}{*{5}{C}}
    LQ & PULSE & DFDNet & PSFR-GAN (ours) & Parsing Map (ours)
    \end{tabularx}
  \caption{Results of $5$-th Solvay conference test.}
  \label{fig:solvay-test1}
\end{figure*}

\begin{figure*}[h]
  \centering
  \includegraphics[width=.99\textwidth]{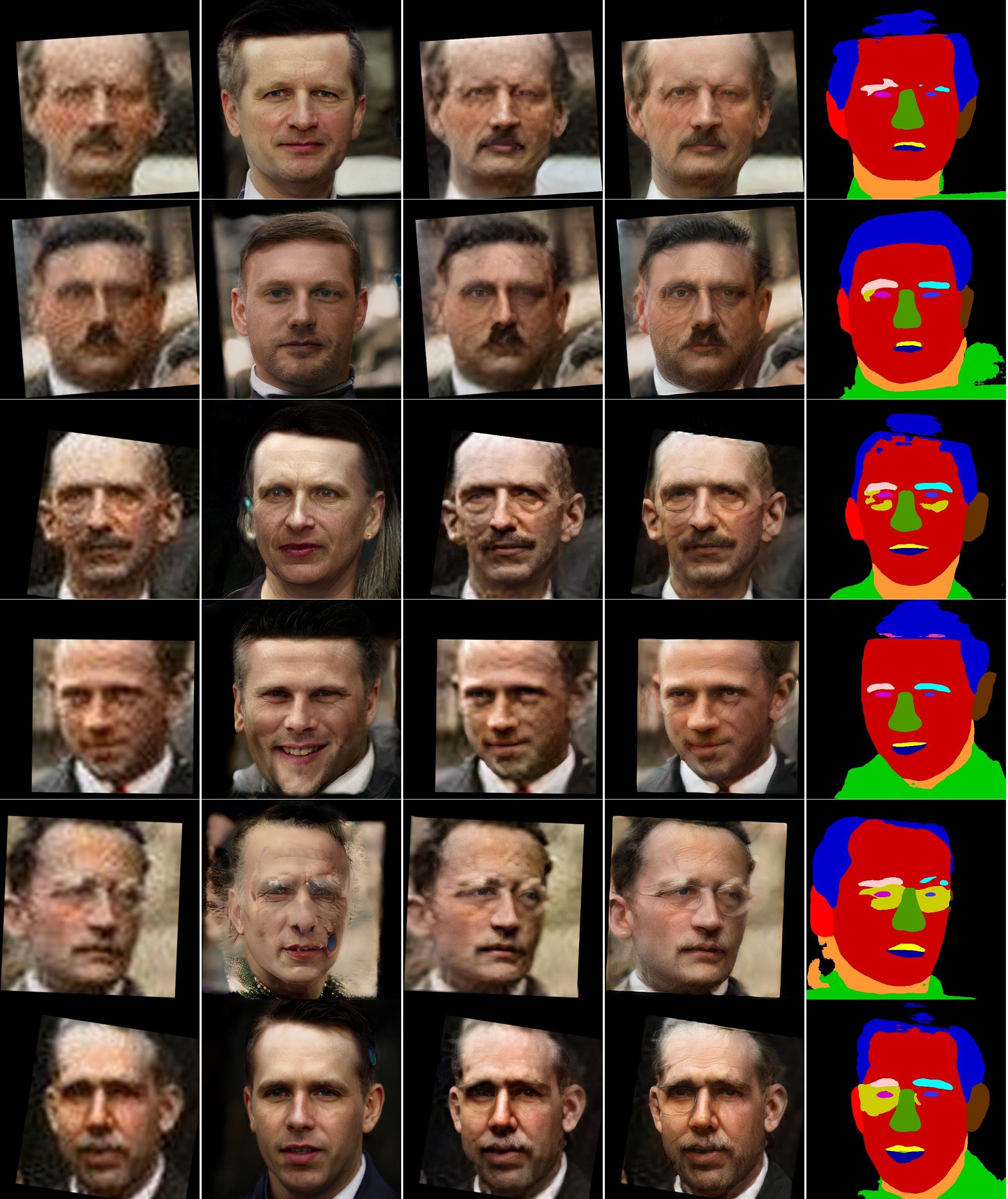}
  \begin{tabularx}{.99\linewidth}{*{5}{C}}
    LQ & PULSE & DFDNet & PSFR-GAN (ours) & Parsing Map (ours)
    \end{tabularx}
  \caption{Results of $5$-th Solvay conference test.}
  \label{fig:solvay-test2}
\end{figure*}

\begin{figure*}[h]
  \centering
  \includegraphics[width=.99\textwidth]{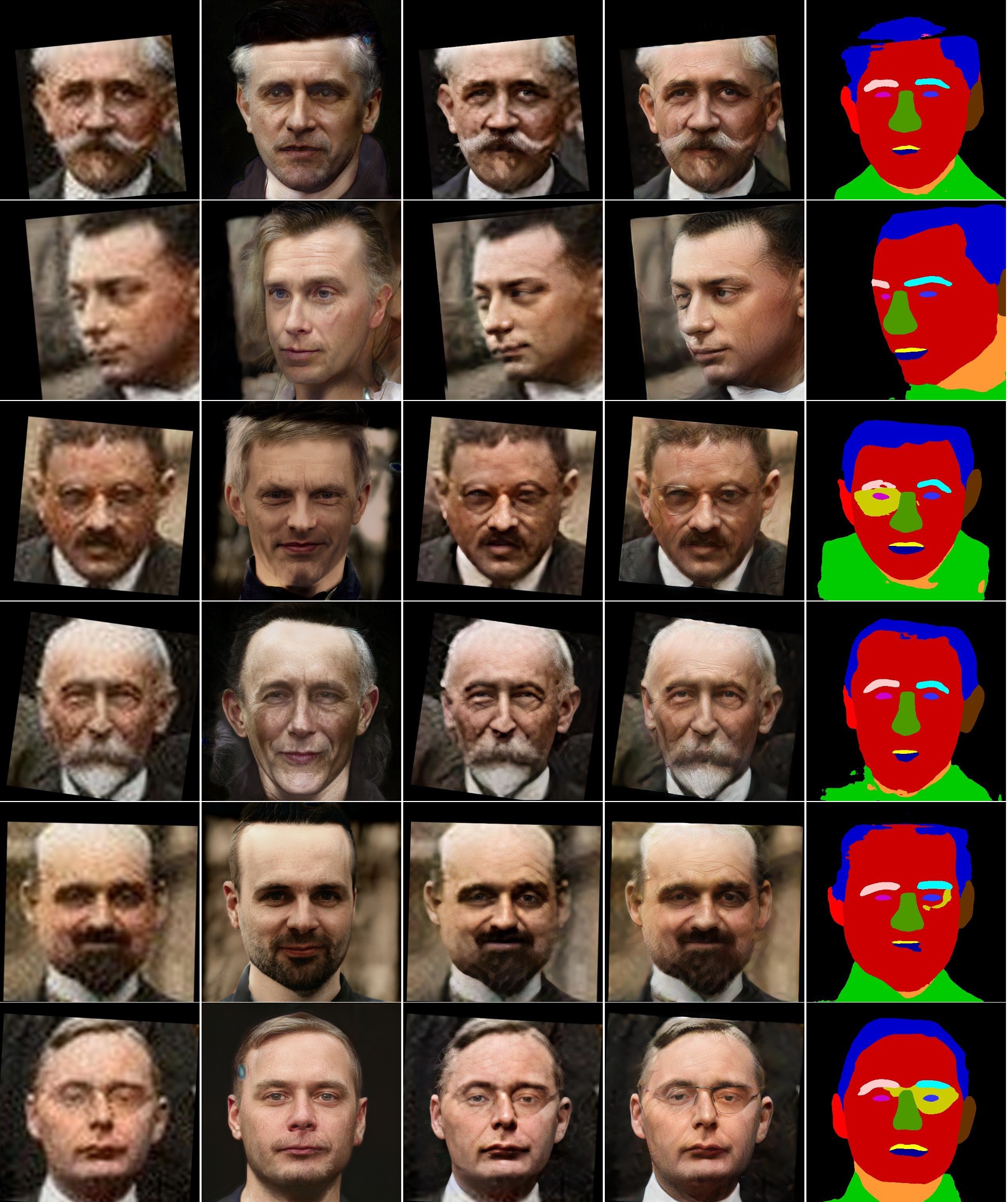}
  \begin{tabularx}{.99\linewidth}{*{5}{C}}
    LQ & PULSE & DFDNet & PSFR-GAN (ours) & Parsing Map (ours)
    \end{tabularx}
  \caption{Results of $5$-th Solvay conference test.}
  \label{fig:solvay-test3}
\end{figure*}

\begin{figure*}[h]
  \centering
  \includegraphics[width=.99\textwidth]{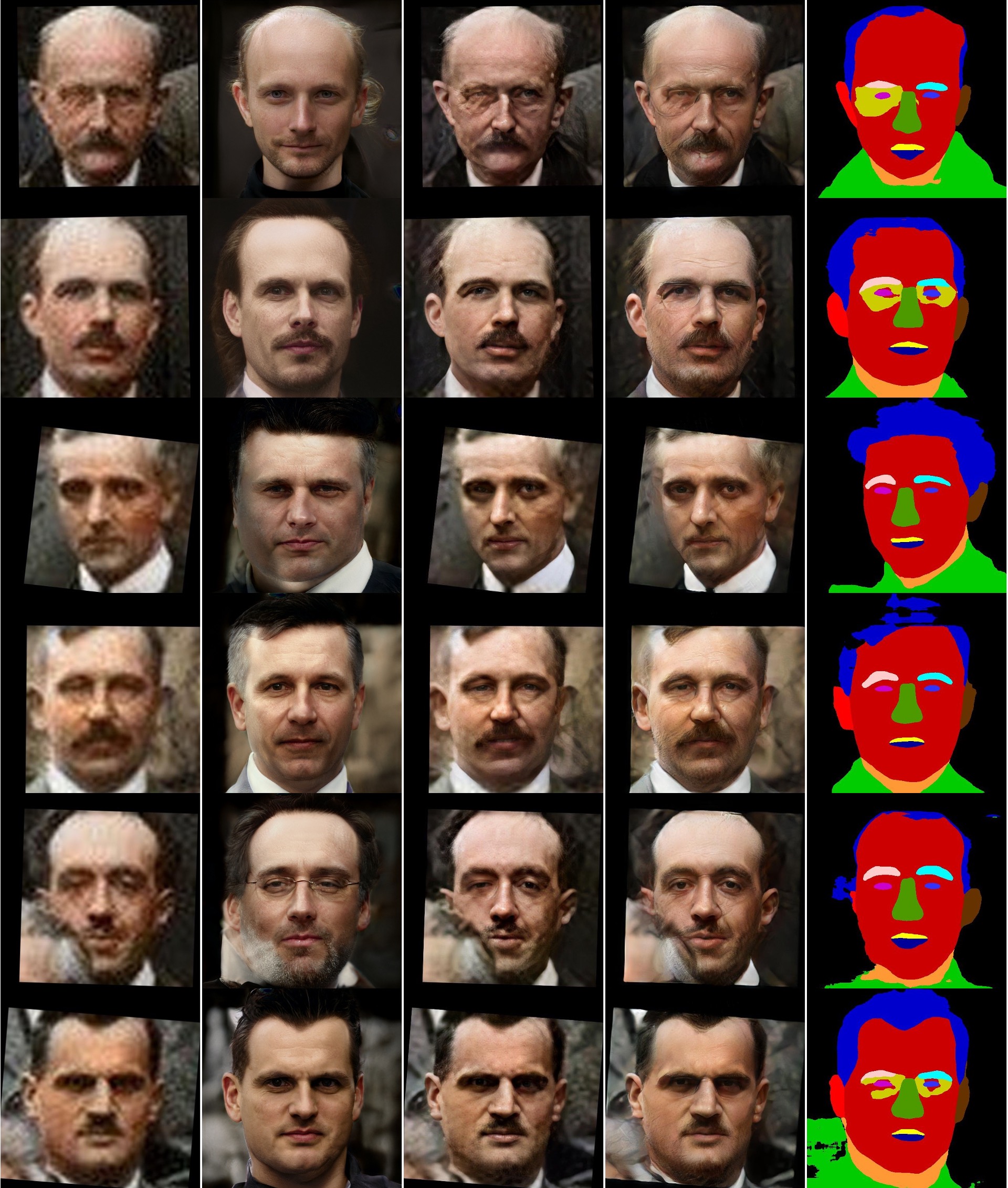}
  \begin{tabularx}{.99\linewidth}{*{5}{C}}
    LQ & PULSE & DFDNet & PSFR-GAN (ours) & Parsing Map (ours)
    \end{tabularx}
  \caption{Results of $5$-th Solvay conference test.}
  \label{fig:solvay-test4}
\end{figure*}

\begin{figure*}[h]
  \centering
  \includegraphics[width=.99\textwidth]{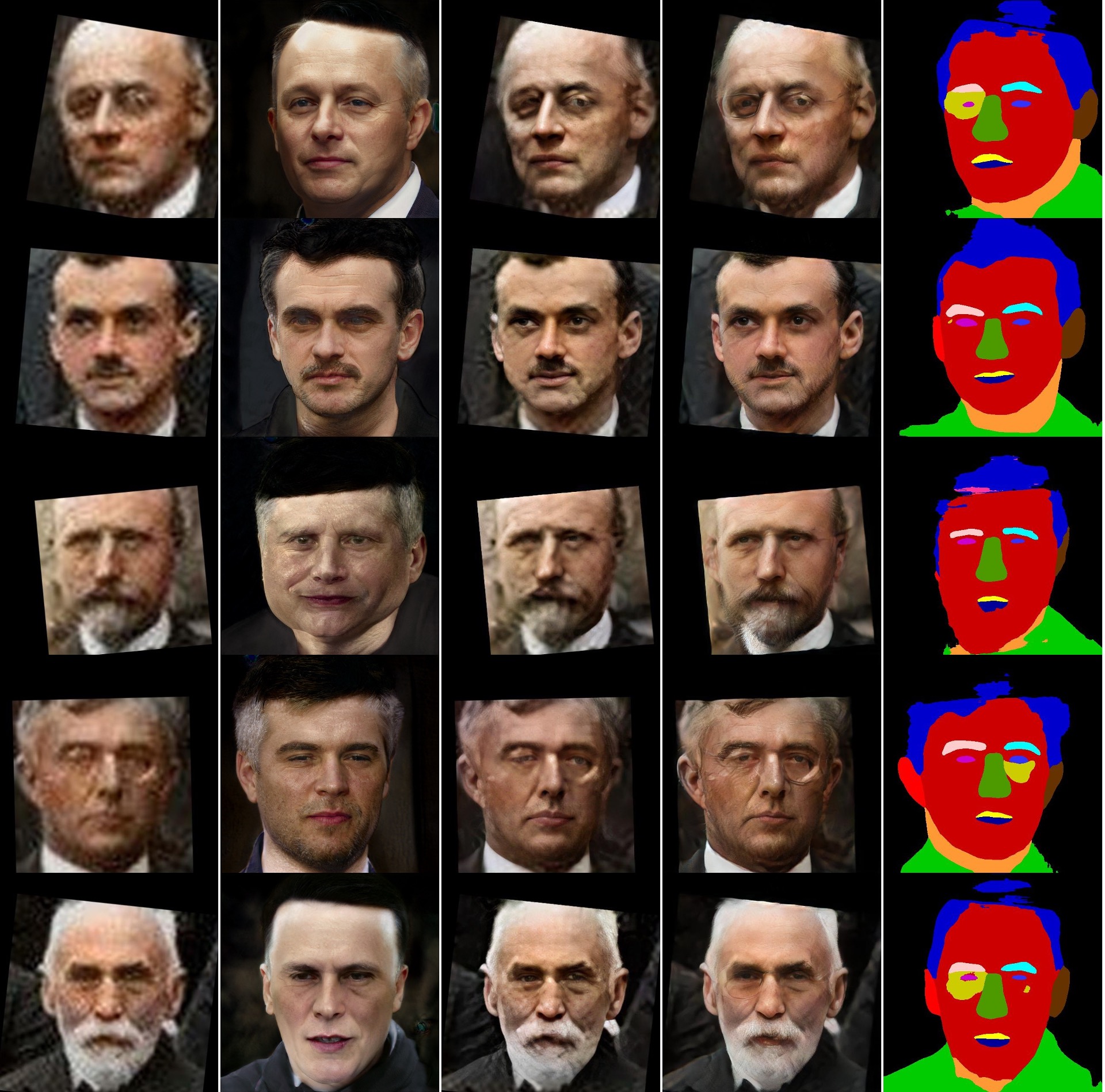}
  \begin{tabularx}{.99\linewidth}{*{5}{C}}
    LQ & PULSE & DFDNet & PSFR-GAN (ours) & Parsing Map (ours)
    \end{tabularx}
  \caption{Results of $5$-th Solvay conference test.}
  \label{fig:solvay-test5}
\end{figure*}

We give the overall results of the $5$-th Solvay conference test images in Fig. \ref{fig:solvay-test-overall}. All faces are cropped out and aligned first, then enhanced by our model and finally paste back to the original photo. Complete results and detailed comparison with other methods are presented in Fig. \ref{fig:solvay-test1}, Fig. \ref{fig:solvay-test2}, Fig. \ref{fig:solvay-test3}, Fig. \ref{fig:solvay-test4} and Fig. \ref{fig:solvay-test5}.

\end{document}